\def\year{2022}\relax
\documentclass[letterpaper]{article} 
\usepackage{aaai22}  
\usepackage{times}  
\usepackage{helvet}  
\usepackage{courier}  
\usepackage[hyphens]{url}  
\usepackage{graphicx} 
\usepackage{amsmath}
\usepackage{multicol}
\usepackage{multirow}
\usepackage{amssymb}
\usepackage{natbib}  
\usepackage{caption} 
\DeclareCaptionStyle{ruled}{labelfont=normalfont,labelsep=colon,strut=off} 
\usepackage{algorithm}
\usepackage{algorithmic}

\usepackage{newfloat}
\usepackage{listings}
\floatstyle{ruled}
\newfloat{listing}{tb}{lst}{}
\floatname{listing}{Listing}
\title{VRAG: Region Attention Graphs for Content-Based Video Retrieval}
\author {
    Kennard Ng,\textsuperscript{\rm 1}
    Ser-Nam Lim, \textsuperscript{\rm 2}
    Gim Hee Lee \textsuperscript{\rm 1}
}
\affiliations {
    \textsuperscript{\rm 1} Department of Computer Science, National University of Singapore \\
    \textsuperscript{\rm 2} Facebook AI\\
    kennard.ng.pool.hua@gmail.com, sernam@gmail.com, gimhee.lee@nus.edu.sg
}

\usepackage{bibentry}

\begin{document}

\maketitle

\begin{abstract}
Content-based Video Retrieval (CBVR) is used on media-sharing platforms for applications such as video recommendation and filtering. To manage databases that scale to billions of videos, video-level approaches that use fixed-size embeddings are preferred due to their efficiency.  In this paper, we introduce Video Region Attention Graph Networks (VRAG) that improves the state-of-the-art of video-level methods. We represent videos at a finer granularity via region-level features and encode video spatio-temporal dynamics through region-level relations. Our VRAG captures the relationships between regions based on their semantic content via self-attention and the permutation invariant aggregation of Graph Convolution. 
In addition, we show that the performance gap between 
video-level and frame-level methods can be reduced by segmenting videos into shots and using shot embeddings for video retrieval. We evaluate our VRAG over several video retrieval tasks and achieve a new state-of-the-art for video-level retrieval. Furthermore, our shot-level VRAG shows higher retrieval precision than other existing video-level methods, and closer performance to frame-level methods at faster evaluation speeds. Finally, our code will be made publicly available.
\end{abstract}

\section{Introduction}

\begin{figure}[t]
    \centering
    \includegraphics[width=0.8\columnwidth]{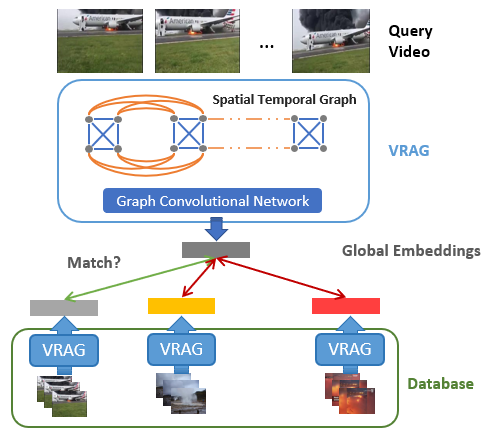}
    \caption{Our VRAG encodes video-level embeddings from spatial intra-frame and temporal inter-frame information for CBVR.} \vspace{-3mm}
    \label{fig:teaser}
\end{figure}

The volume of videos on the Internet has grown exponentially with the inception of media-sharing websites such as Facebook, Twitch and Youtube. 
Content-based Video Retrieval (CBVR) is important on these platforms for applications such as video recommendation and video filtering.

In CBVR, evaluating the video similarity is a key component. There are predominantly two types of approaches for inferring video-to-video similarity: frame-level approaches and video-level approaches. Video-level methods encode videos into fixed size embeddings, and measure video similarity through the similarity of their embeddings. On the other hand, frame-level methods derive video similarity by aggregating pairwise similarities between video frames. These two approaches form a dichotomy. Video-level approaches offer faster evaluation speeds, while frame-level methods provide more accurate video similarities at significant overheads. Although practical implementations of frame-level methods may employ optimizations such as summarizing the video into a shorter sequence of frames~\cite{vid-summary:1708-09545, vid-summary:1812-01969, vid-summary:zhou2018deep}, video-level methods remain orders of magnitude more efficient than frame-level methods when the number of videos scale to the size of billions. Consequently, video-level methods remain highly relevant for real-world applications despite its inferior performance to frame-level methods.

Recently, several video-level CBVR works~\cite{lbow, Baraldi2018LAMVLT, kordopatis2017dml, baseline:tmk} propose the same design principle of extracting local frame-level features, and then aggregating these features into fixed-size video embeddings. As a result, most of these works focus on proposing a better frame-level feature extractor and/or video-level feature pooling methods. For example, DML~\cite{kordopatis2017dml} and LBoW~\cite{lbow} extract Maximum Activation of Convolution (MAC) features from each frame and then apply mean pooling and bag-of-words, respectively, to get video-level representations. LAMV~\cite{Baraldi2018LAMVLT} later propose representing frames at a finer granularity using Region Maximum Activation of Convolutions (R-MAC) and learnable pooling of these frame descriptors in the Fourier Domain~\cite{baseline:tmk}. While prior works have made strides in frame-level feature extraction and video-level pooling, these approaches encode each frame separately, and do not model the spatial and temporal interactions inherent within videos. 

\begin{figure}[t]
    \centering
    \includegraphics[width=\columnwidth]{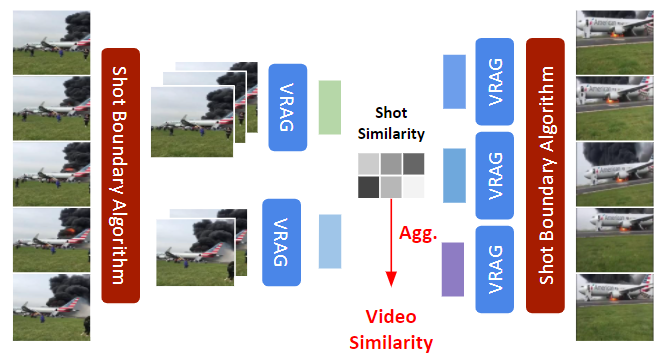}
    \caption{Our VRAG in a shot-level video retrieval setting.} \vspace{-3mm}
    \label{fig:shot-retrieval-diagram}
\end{figure}

In view that prior video-level approaches lack modelling of spatio-temporal interactions within videos, we propose VRAG: a region attention graph-based framework for content-based video retrieval. As shown in Figure~\ref{fig:teaser}, our VRAG models videos at the fine-grained region-level as a graph. Each node of the graph represents a R-MAC feature vector. To model interactions between region nodes, we encode spatial relations through complete subgraphs between regions in the same frame and temporal relations through fully connected edges across adjacent frames. We further augment these spatial and temporal connections through self-attention, which modulates the strength of these associations through the affinities between their region-level content. Consequently, we transform each region into context-aware embeddings by selectively aggregating features from neighboring regions via Graph Attention~\cite{graph-attention} layers. To generate video-level embeddings, we learn attention weights for each region and selectively aggregate region-level embeddings into a fixed size representation. We model our attention pooling on the intuition that important video regions add significant context to other regions and derive the attention weights from the pairwise affinities between regions across multiple Graph Attention layers. 

Our VRAG improves the state-of-the-art for video-level approaches across video retrieval tasks~\cite{dataset:fivr200k, dataset:evve, dataset:cc-web-video}, and reduces the performance gap between video-level and frame-level retrieval approaches. On Event Video Retrieval~\cite{dataset:evve}, our VRAG also achieves higher retrieval scores over the state-of-the-art for frame-level retrieval, ViSiL~\cite{kordopatiszilos2019visil}, while being more than 50$\times$ faster. We further propose to reduce the gap between video-level and frame-level approaches by segmenting videos into shots and representing videos over multiple shot embeddings using our VRAG, as illustrated in Figure~\ref{fig:shot-retrieval-diagram}. Our shot-level VRAG evaluates the similarity between videos by aggregating pairwise similarities between their shot embeddings. In our experiments, we show that our shot-level VRAG bridges the gap between video-level and frame-level approaches on most video retrieval tasks~\cite{dataset:fivr200k, dataset:cc-web-video}
at faster evaluation speeds than frame-level methods, and higher retrieval precision over video-level approaches. 

\section{Related Work}

\subsection{Video Retrieval}

\textbf{Multi-modal video retrieval} performs video retrieval through multi-modal embeddings. Recent methods use pretrained expert networks to extract video representations across different modalities, e.g. optical flow, audio, and introduce novel techniques to fuse these representations into a multi-modal embedding~\cite{collaborative-experts, moEE, mm-transformer}. \cite{collaborative-experts} models pairwise interactions between modalities and aggregates across modalities via attention pooling while \cite{mm-transformer} use Transformers~\cite{attention-is-all-you-need} to fuse features from different modalities.

\textbf{Video-to-text retrieval} is a popular instance of cross-modal video retrieval, where relevant videos are retrieved given a query sentence and vice versa. The video-to-text pipeline extracts video and text representations and map related video-text pairs into the same representation~\cite{collaborative-experts, howto100m}.

\subsection{Content-based Video Retrieval (CBVR)}
CBVR methods can be broadly classified into video-level~\cite{baseline:tmk, Baraldi2018LAMVLT, kordopatis2017dml, lbow} and frame-level~\cite{temporal_network, baseline:dp, kordopatiszilos2019visil} approaches.

\textbf{Video-level approaches} encode videos into fixed size embeddings, i.e.  vectors. These methods extract features from video frames and temporally pool frame features into a fixed size representation. The similarity between videos is then determined through the similarities between their embeddings e.g. cosine similarity. Compared to prior video-level~\cite{kordopatis2017dml, lbow, baseline:tmk, Baraldi2018LAMVLT} methods that encode video frames independently, our VRAG models spatio-temporal relations between/within frames through Graph Convolution, and adds video-level context into the frame representations.

\textbf{Frame-level approaches}~\cite{temporal_network, baseline:dp} represent videos as sequences of video frame features that scale with the video length. To evaluate video similarity, these methods aggregate pairwise similarities between video frames to derive the similarity between videos. Recently, ViSiL~\cite{kordopatiszilos2019visil} propose using Convolutional Neural Networks (CNN) layers and the (Symmetric) Chamfer Similarity to aggregate pairwise frame similarities, and achieves state-of-the-art frame-level CBVR. However, ViSiL requires $n^2$ forward passes to evaluate pairwise similarities between $n$ videos, while our VRAG is orders of magnitude faster since it uses only $n$ forward passes to generate video embeddings.

\section{Our Approach}
\begin{figure*}[t]
    \centering
    \includegraphics[width=0.85\textwidth]{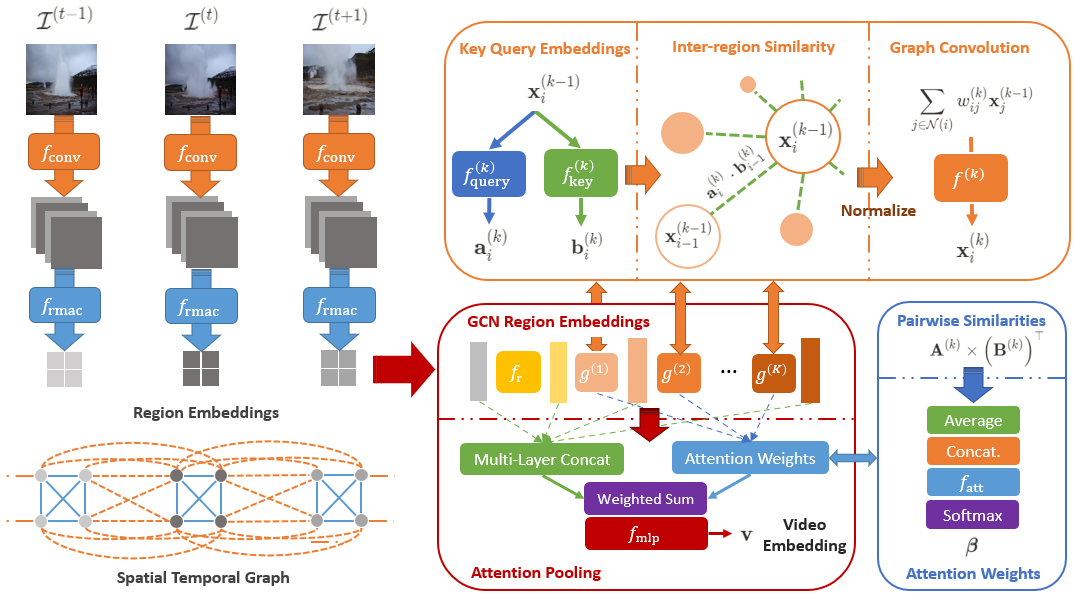}
    \caption{Our Video Region Attention Graph Network (VRAG). (Left) Graph structure. (Right) VRAG Network.}
    \label{fig:VRAG-architecture}
\end{figure*}

We present our Video Region Attention Graph (VRAG) network in Figure~\ref{fig:VRAG-architecture}. 

\textbf{\label{sect:frame-repr}Input Frame Representation.} Following ViSiL, we represent each video frame using concatenated Region Maximum Activation of Convolution~\cite{rmac} (R-MAC) features from intermediate convolution layers. We pass each video frame through a pretrained ConvNet, i.e. ResNet50~\cite{resnet} pre-trained on ImageNet, with $L$ layers to generate intermediate activation maps, $f_\text{conv}: \mathcal{I}^{(t)} \mapsto \left\{\mathcal{M}^{(t)}_1, \ldots, \mathcal{M}^{(t)}_L\right\}$. Given that each intermediate activation map $\mathcal{M}_l^{(t)} \in \mathbb{R}^{H_l \times W_l \times C_l}$ has varying spatial dimensions, we transform the activation maps into the same spatial dimension by defining different R-MAC kernel sizes for each layer. Subsequently, we concatenate the intermediate R-MAC features channel-wise to represent each video frame, i.e. $f_\text{rmac}: \left\{\mathcal{M}^{(t)}_1, \ldots, \mathcal{M}^{(t)}_L\right\} \mapsto \mathcal{X}^{(t)} \in \mathbb{R}^{R \times C}$, where $R$ refers to the number of R-MAC features in each frame, and 
$C=\sum_{l=1}^{L}C_l$ 
is the dimensions of the concatenated vector. We then encode a video of length $T$ into $N=T\times R$ R-MAC features, i.e. $\mathbf{X} = \left[\mathcal{X}^{(1)}|\mathcal{X}^{(2)}| \ldots| \mathcal{X}^{(T)}\right] \in \mathbb{R}^{N \times C}$.

\textbf{Spatial and Temporal Graph Structure.}
We represent each R-MAC feature vector $\mathbf{x}_i \in \mathbf{X}=\left[\mathbf{x}_1, \ldots, \mathbf{x}_N\right] \in \mathbb{R}^{N \times C}$ as a node $\mathcal{V}_i$ in the graph $\mathcal{G} = \{\mathcal{V}, \mathcal{E}_\text{spatial}, \mathcal{E}_\text{temporal}\}$ of our VRAG.
We then add two sets graph edges: 1) $\mathcal{E}_\text{spatial}$ is a set of edges that connects region nodes in the same frame $\mathcal{I}^{(t)}$ to capture spatial relationships between these nodes with a frame and this includes a self referencing edge. 2) $\mathcal{E}_\text{temporal}$ refers to edges that connect the regions between consecutive frames $\mathcal{I}^{(t-1)}$ and $\mathcal{I}^{(t+1)}$ to capture temporal relations between video frames. 

\textbf{Learning Region Embeddings.} From the video region graph $\mathcal{G} = \{\mathcal{V}, \mathcal{E}_\text{spatial}, \mathcal{E}_\text{temporal}\}$ constructed earlier, we learn region-level embeddings using Graph Convolutional Network~\cite{gcn, graph-attention} (GCN) layers. To increase the duration of videos that can be processed, we first reduce the dimensions of R-MAC region vectors, i.e. $\mathbf{X} \in \mathbb{R}^{N\times C}$ using a fully-connected layer with non-linear activation, i.e. $f_{r}: \mathbf{X} \mapsto \mathbf{X}^{(0)} \in \mathbb{R}^{N\times C^\prime}$.
We then pass the region vectors through $K$ Graph Attention~\cite{graph-attention} layers to generate intermediate region features, i.e. $g^{(k)}: \mathbf{X}^{(k-1)} \mapsto \mathbf{X}^{(k)} \in \mathbb{R}^{N \times C^\prime}$ for $k \in \left[1, \ldots, K\right]$. Specifically, we aggregate the features from neighboring regions as follows: 

\paragraph{1)} To enable interactions between neighboring regions with complementary representations, we use the key-query self-attention mechanism~\cite{attention-is-all-you-need} to describe the similarity between regions. Specifically, we generate key and query embeddings using linear transformations from the input region vectors $\mathbf{x}_i^{(k-1)}$, i.e.:
\begin{equation}
\small
        f_\text{query}^{(k)} : \mathbf{x}_i^{(k-1)} \mapsto \mathbf{a}_i^{(k)} \in \mathbb{R}^{C^\prime}, \quad
        f_\text{key}^{(k)} : \mathbf{x}_i^{(k-1)} \mapsto \mathbf{b}_i^{(k)} \in \mathbb{R}^{C^\prime}.
    \label{eqn:kq-embeddings}
\end{equation}

\vspace{-2mm}
\paragraph{2)} The similarity between a region $i$ and its neighboring region $j \in \mathcal{N}(i)$ is defined using the dot product of their query and key embeddings, i.e. $s_{ij}^{(k)}= \mathbf{a}_i^{(k)} \cdot \mathbf{b}_{j}^{(k)}$. The output of the Graph Attention layer $g^{(k)}$ is derived as:
\begin{equation}
       w_{ij}^{(k)} = \frac{e^{s_{ij}^{(k)}}}{\sum_{n \in \mathcal{N}(i)}{e^{s_{in}^{(k)}}}}, \quad
       f^{(k)} : \sum_{j \in \mathcal{N}(i)} w_{ij}^{(k)} \mathbf{x}_{j}^{(k-1)} \mapsto \mathbf{x}_i^{(k)},
\end{equation}
where $f^{(k)}$ is a linear layer with a non-linear activation.

\vspace{-2mm}
\paragraph{3)} Finally, given that adding graph convolution layers may lead to over-smoothing of region representations~\cite{gcn:over-smoothing1, chen2019measuring}, we propose concatenating region representations along the depth of our network to preserve discriminative features, i.e. $f_\text{concat}: \left\{\mathbf{X}, \mathbf{X}^{(0)}, \ldots, \mathbf{X}^{(K)}\right\} \rightarrow \mathbf{R} \in \mathbb{R}^{N \times \left(C + (K+1)C^\prime\right)}$.

\textbf{Video Embedding.} We aggregate region embeddings $\mathbf{R}=[\mathbf{r}_1, \ldots, \mathbf{r}_N] \in \mathbb{R}^{N \times \left(C + (K+1)C^\prime\right)}$ using weighted sum aggregation to generate a pooled region representation:
    $\overline{\mathbf{r}}=\sum^{N}_{i=1}\beta_i \mathbf{r}_i$,
where $\beta_i$ is the attention weight for region $\mathbf{r}_i$. The pooled representation $\overline{\mathbf{r}}$ is fed into two linear layers to generate the video-level embedding $\mathbf{v}$, i.e.: $f_\text{mlp}:\overline{\mathbf{r}} \mapsto \mathbf{v} \in \mathbb{R}^D$. We model our attention weights $\boldsymbol{\beta} = [\beta_1, \ldots, \beta_N] \in \mathbb{R}^N$ from the intuition that important regions in a video act as anchor regions that add context to other regions, and should have high affinities with other regions in the video. Therefore, we compute the attention weights $\boldsymbol{\beta}$ as follows:
\vspace{-2mm}
\paragraph{1)} We reuse the key and query embeddings of each Graph Attention layer $g^{(k)}$ from Equation~\ref{eqn:kq-embeddings} to compute $K$ pairwise affinity matrices $\mathcal{A}^{(k)}$ between regions, i.e.
\begin{equation}
    \begin{aligned}
    \mathcal{A}^{(k)}&=\mathbf{A}^{(k)} \times \left(\mathbf{B}^{(k)}\right)^\top \in \mathbb{R}^{N \times N}, \\
    \mathbf{A}^{(k)} &= [\mathbf{a}_i^{(k)}, \ldots, \mathbf{a}^{(k)}_N] \in \mathbb{R}^{N \times C^\prime}, \\
    \mathbf{B}^{(k)} &= [\mathbf{b}_i^{(k)}, \ldots, \mathbf{b}^{(k)}_N] \in \mathbb{R}^{N \times C^\prime}. \\
    \end{aligned}
\end{equation}
The row $\mathcal{A}_i^{(k)}$ returns the pairwise affinity between region $i$ and every other region $j \in [1, \ldots, N]$ at Graph Attention layer $k$. For each region $i$, we average the affinities in $\mathcal{A}_i^{(k)}$ and concatenate the averaged affinities over $K$ Graph Attention layers to obtain the affinity vector $\boldsymbol{\alpha}_i \in \mathbb{R}^{K}$. 

\vspace{-2mm}
\paragraph{2)} We feed $\boldsymbol{\alpha}_i$ into a single linear layer to derive the unnormalized attention weights i.e. $f_\text{att}: \boldsymbol{\alpha}_i \mapsto \tilde{\beta}_i$ and normalize the region attention weights 
via softmax, i.e.:
\begin{equation}
   \beta_i = \frac{e^{\tilde\beta_i}}{\sum^{N}_{j=1}e^{\tilde\beta_j}}.
\end{equation}

\subsection{Training VRAG}
We train VRAG using the triplet margin loss~\cite{triplet-margin-loss}. From a triplet of video-level embeddings $\left(\mathbf{v}, \mathbf{v}^+, \mathbf{v}^-\right)$ that correspond to the anchor, positive and negative video respectively, we compute the loss:
\begin{equation}
    \mathcal{L} = \left\lfloor c\left(\mathbf{v}, \mathbf{v}^-\right) - c\left(\mathbf{v}, \mathbf{v}^{+}\right) + m\right\rfloor_+,
\end{equation}
where $c(\cdot, \cdot)$ computes the cosine similarity between embeddings, $m$ refers to the margin. During training, we sample triplets using the triplet mining scheme consistent with ~\cite{kordopatiszilos2019visil}.

\section{Shot Representations}

In this section, we introduce an intermediate approach utilizing shot representations for video retrieval. We use the shot-boundary algorithm in Section~\ref{sect:sba} to divide a video $\left[\mathcal{I}^{(1)}, \ldots, \mathcal{I}^{(T)}\right]$ into non-overlapping shots and use VRAG to encode each shot into embeddings. Using shots can reduce noise within VRAG embeddings by removing spurious edges across shot boundaries.
In our shot-level approach, we evaluate video similarity by aggregating the pairwise similarities between shot-level embeddings.

\textbf{\label{sect:sba}Shot Boundary Algorithm.} In our shot boundary algorithm, we identify the boundaries between video shots by comparing the similarities between consecutive frames. Specifically, we flatten the R-MAC features from each frame $\mathcal{X}^{(t)} \in \mathbb{R}^{R \times C}$ into the frame representation vector $\mathcal{F}^{(t)} \in \mathbb{R}^{(R\times C)\times1}$. The similarities between consecutive frames is then the cosine similarity between their representations vectors $\mathcal{F}^{(t)}$. We mark the frame $t$ as the start of a new video shot when the cosine similarity between $\mathcal{F}^{(t)}$ and $\mathcal{F}^{(t-1)}$ is lower than the minimum cosine similarity threshold $\tau_s$. Generally, a higher $\tau_s$ creates more shots and resembles frame-level approaches while a smaller $\tau_s$ reduces the number of shots and closely approximates video-level methods.

\textbf{\label{sect:aggregate-shots}Aggregating Shot Similarities.} Under our shot-level approach, we evaluate the similarity between videos $\mathbf{X}$ and $\mathbf{X}^\prime$ from the pairwise similarities between their shot embeddings, i.e. $\left[\mathbf{o}_1, \ldots, \mathbf{o}_N\right]$ and $\left[\mathbf{o}_1^\prime, \ldots, \mathbf{o}_M^\prime\right]$. We compute the cosine similarity between pairs of shots and build a pairwise shot cosine similarity matrix $\mathcal{S} \in [-1, 1]^{N \times M}$ where $\mathcal{S}_{ij}$ returns the cosine similarity between shots $\mathbf{o}_i$ and $\mathbf{o}_j^\prime$.

We aggregate these pairwise similarities into a similarity metric between videos. We use two aggregation schemes proposed in \cite{kordopatiszilos2019visil}: Chamfer similarity (CS) which takes the maximum similarity along the columns of $\mathcal{S}$ followed by average pooling of the maximum similarity vector, i.e. 
\begin{equation}
    s = \operatorname{CS}(\mathbf{X}, \mathbf{X}^\prime)=\frac{1}{N}\sum_{i=1}^N \left(\max_{j \in 1, \ldots, M} \mathcal{S}_{ij}  \right),
\end{equation}
Symmetric Chamfer Similarity (SCS) which computes the average Chamfer similarity from $\mathcal{S}$ and $\mathcal{S^\top}$, i.e.:
\begin{equation}
    \begin{aligned}
    s = \operatorname{SCS}(\mathbf{X}, \mathbf{X}^\prime) = \frac{\operatorname{CS}(\mathbf{X}, \mathbf{X}^\prime) + \operatorname{CS}(\mathbf{X}^\prime, \mathbf{X})}{2}.
    \end{aligned}
\end{equation}

\section{Experiments}

We evaluate video retrieval performance using mean Average Precision~\cite{dataset:cc-web-video} (mAP). Additionally, we provide qualitative results in our supplementary material.

\subsection{Experiment Settings}

We evaluate VRAG and our shot-level approach over several video retrieval settings: Near-Duplicate Video Retrieval (NDVR); Event Video Retrieval (EVR); and Fine-grained Incident Video Retrieval (FIVR).

\vspace{0mm}
\paragraph{Datasets.} For evaluation, we use CC\_WEB\_VIDEO~\cite{dataset:cc-web-video} for NDVR, EVVE~\cite{dataset:evve} for EVR, and FIVR200K~\cite{dataset:fivr200k} and its subset FIVR5K~\cite{dataset:fivr200k, kordopatiszilos2019visil} for FIVR. During training, we sample triplets of videos from VCDB~\cite{dataset:vcdb}.

\noindent \textbf{CC\_WEB\_VIDEO}~\cite{dataset:cc-web-video} contains 13,129 videos with 24 query videos. In this dataset, near-duplicate videos correspond to videos stored in different formats e.g. .flv; .wmv; and videos with minor content differences e.g. photometric variations like lighting. We evaluate on the original annotations from \cite{dataset:cc-web-video} and the cleaned annotations from \cite{kordopatiszilos2019visil}. 

\noindent\textbf{EVVE}~\cite{dataset:evve} contains 2,995 videos from Youtube with 620 query videos over 13 events. 

\noindent \textbf{FIVR200K}~\cite{dataset:fivr200k} contains 100 query videos and a total of 225,960 videos from 4,687 Wikipedia events. The dataset groups similar videos into four categories: 1) Near-Duplicate (ND) videos which share all scenes with the query video; 2) Duplicate Scene (DS) videos that share at least one scene with the query video; 3) Complementary Scene (CS) videos that share at least segment with the query video, but from a different viewpoint; 4) Incident Scene (IS) videos that capture the same incident as the query video without sharing segments.
From the video categories, the dataset introduces three video retrieval settings: 1) Duplicate Scene Video Retrieval (DSVR) which only considers ND and DS videos as positives; 2) Complementary Scene Video Retrieval (CSVR) that extends from DSVR to include CS videos; and 3) Incident Scene Video Retrieval (ISVR) which marks all four categories of videos as similar videos.

\vspace{0mm}
\noindent \textbf{FIVR5K}~\cite{dataset:fivr200k, kordopatiszilos2019visil} is a subset of FIVR200K, with 50 query videos and 5,000 database videos. To create FIVR5K, the authors~\cite{kordopatiszilos2019visil} select the 50 hardest query videos on the DSVR task from FIVR200K using \cite{lbow} to measure difficulty.

\vspace{0mm}
\noindent \textbf{VCDB}~\cite{dataset:vcdb} consists of a core dataset with 528 videos and a background dataset with 100,000 distractor videos. The core dataset videos has a total of 9,236 pairs of partial copies between the 528 videos in the core dataset.

\paragraph{Implementation Details.}
For each video, we sample frames at one second intervals. We extract R-MAC features from each bottleneck layer of ResNet50~\cite{resnet}, i.e. $L=4$ and $C=3840$. We set the number of regions per frame $R=9$, intermediate dimensions $C^\prime=512$, video-level embedding dimensions $D=4096$. We use $K=3$ Graph Attention~\cite{graph-attention} layers and the ELU~\cite{nonlinearity:elu} non-linearity.
We train VRAG on a single GTX 1080 Ti using a batch size of one over 120 epochs. The maximum duration of each video clip in the training triplet is $W=64$s, and we sample 1,000 triplets from each triplet pool, giving a total of 2,000 training iterations per epoch. We optimize VRAG using the triplet margin loss with $m=0.2$ and the Adam~\cite{optimizer:adam} optimizer using a fixed learning rate at $3 \times 10^{-7}$. During inference, VRAG processes up to 2,000 frames on 11GB of GPU memory, which amounts to over 30 minutes of video.

\subsection{Ablation Studies}



\paragraph{Our VRAG.}
At the region-level, we use attention to aggregate features from neighboring regions. We compare our attention aggregation mechanism with alternatives such as max aggregation and average aggregation, i.e. GCN~\cite{gcn}. In Table~\ref{tab:ablation:region-aggregation}, we show that our choice of region-level attention aggregation improves performance across all FIVR5K settings.

\begin{table}[ht]
    \centering
    \begin{tabular}{|c|c|c|c|c|}
    \hline
         Method & Region Agg. & DSVR & CSVR & ISVR \\
         \hline\hline
         \multirow{3}{*}{VRAG} & Max & 0.520 & 0.537 & 0.508 \\
         \cline{2-5}
         &  Average & 0.518 & 0.532 & 0.506 \\
         \cline{2-5}
          &  Attention & \textbf{0.532} & \textbf{0.548} & \textbf{0.518} \\
         \hline
    \end{tabular}
    \caption{Comparison on FIVR5K over different choices of region-level aggregation. Attention-weighted summation is used to pool region embeddings for all choices of region-level aggregation.}
    \label{tab:ablation:region-aggregation}
\end{table}

At the video-level, we pool region embeddings to a fixed size vector $\bar{\mathbf{r}}$ using an attention-weighted summation. We compare our attention pooling with standard methods, i.e. max pooling and average pooling. In Table~\ref{tab:ablation:region-pooling}, we demonstrate that our learnable pooling gives better performance over conventional pooling techniques.  

\begin{table}[ht]
    \centering
    \begin{tabular}{|c|c|c|c|c|}
    \hline
         Method & Region Pooling & DSVR & CSVR & ISVR \\
         \hline\hline
         \multirow{3}{*}{VRAG} & Max & 0.423 & 0.435 & 0.415 \\
        \cline{2-5}
          & Average & 0.505 & 0.513 & 0.499 \\
         \cline{2-5}
          &  Attention & \textbf{0.532} & \textbf{0.548} & \textbf{0.518} \\
         \hline
    \end{tabular}
    \caption{\label{tab:ablation:region-pooling} {Comparison on FIVR5K over different video-level pooling techniques.}}
\end{table}
We also compare our choice of self-attention to an alternative with reduced parameters. In Table~\ref{tab:ablation:attention}, we observe that using separate key/query parameters allows for more complex relations between regions and significantly improves retrieval performance in FIVR5K.
\begin{table}[ht]
    \centering
    \begin{tabular}{|c|c|c|c|c|}
    \hline
         Method & Attention & DSVR & CSVR & ISVR \\
         \hline\hline
         \multirow{2}{*}{VRAG} & $f_\text{query}^{(k)} = f_{key}^{(k)}$ & 0.493 & 0.510 & 0.490 \\
         \cline{2-5}
          &  $f_\text{query}^{(k)} \neq f_{key}^{(k)}$ & \textbf{0.532} & \textbf{0.548} & \textbf{0.518} \\
         \hline
    \end{tabular}
    \caption{Comparison on FIVR5K using different implementations of attention.}
    \label{tab:ablation:attention}
\end{table}

Finally, we propose concatenating the region-level embeddings to reduce over-smoothing effects observed in GCNs~\cite{gcn:over-smoothing1, chen2019measuring}. We compare against three baselines: 1) Using only the output from the final Graph Attention layer; 2) Concatenating the outputs from all Graph Attention layers; 3) Concatenating the outputs from all Graph Attention layers and $f_r$. In Table~\ref{tab:ablation:region-embed}, we show that  multi-layer concatenation of region-level embeddings greatly improves retrieval performance. Furthermore, we observe a monotonic increase in performance as we concatenate  region representations across more layers.

\begin{table}[ht]
    \centering
    \begin{tabular}{|c|c|c|c|c|}
    \hline
         Method & Concat. Layers & DSVR & CSVR & ISVR \\
         \hline\hline
         \multirow{4}{*}{VRAG} & Final Graph Attn. & 0.452 & 0.463 & 0.442 \\
        \cline{2-5}
         & All Graph Attn. & 0.481 & 0.497 & 0.477 \\
        \cline{2-5}
         & All Graph Attn. + $f_r$ & 0.507 & 0.526 & 0.498 \\
        \cline{2-5}
         &  All Layers & \textbf{0.532} & \textbf{0.548} & \textbf{0.518} \\
         \hline
    \end{tabular}
    \caption{Comparison on FIVR5K over different levels of depth-wise concatenation.}
    \label{tab:ablation:region-embed}
\end{table}

\paragraph{Shot-level Video Retrieval.}
In shot-level video retrieval, i.e. VRAG-S, we compare different choices of $\tau_s$ for our shot boundary algorithm. In Table~\ref{tab:ablation:shot-level}, we demonstrate that increasing the number of shots i.e $\tau_s = 0.75$ results in better performance in FIVR5K. Our results are also consistent with the frame-level approach ViSiL~\cite{kordopatiszilos2019visil}, where aggregating similarities using Chamfer Similarity (CS) has better performance than Symmetric Chamfer Similarity (SCS). We also observe that our shot-level approach bridges the large gap in performance between video level approaches, i.e. VRAG and frame-level approaches, i.e. ViSiL~\cite{kordopatiszilos2019visil}, in FIVR5K.


\begin{table}[ht]
    \centering
    \begin{tabular}{|c|c|c|c|c|c|}
    \hline
         Method & $\tau_s$ & Shot Agg. & DSVR & CSVR & ISVR \\  
         \hline\hline
         VRAG & - & - & 0.532 & 0.548 & 0.518 \\
         \hline \hline
         \multirow{4}{*}{VRAG-S} & \multirow{2}{*}{0.5} & CS & 0.493 & 0.512 & 0.483 \\
          \cline{3-6}
         && SCS & 0.463 & 0.477 & 0.464 \\
          \cline{2-6}
         & \multirow{2}{*}{0.75} & CS & \textbf{0.709} & \textbf{0.704} & \textbf{0.636} \\
         \cline{3-6}
        & & SCS & 0.596 & 0.606 & 0.575 \\
         \hline \hline 
         \multirow{2}{*}{ViSiL} & \multirow{2}{*}{-} & CS & \textbf{0.880} & \textbf{0.869} & \textbf{0.777} \\
         \cline{3-6}
         & & SCS & 0.830 & 0.823 & 0.731 \\
         \hline \hline
    \end{tabular}
    \caption{Comparison on FIVR5K over different shot-level hyperparameters.}
    \label{tab:ablation:shot-level}
\end{table}

\subsection{Comparison with Baseline Methods}

We compare VRAG and our shot-level approach with existing approaches as baselines across several video retrieval tasks. We evaluate our performance on CC\_WEB\_VIDEO~\cite{dataset:cc-web-video} for NDVR, EVVE~\cite{dataset:evve} for EVR, and use FIVR200K~\cite{dataset:fivr200k}, and FIVR5K~\cite{kordopatiszilos2019visil, dataset:fivr200k}, for FIVR.
\vspace{0mm}
\paragraph{Video-level Methods.} For LBoW~\cite{lbow}, we use the results from \cite{video-verification-fake-news} for CC\_WEB\_VIDEO and FIVR200K. We re-implement LBoW, i.e. LBoW$^\dagger$, with 1000 codebooks built on  VCDB using KMeans++, for EVVE and FIVR5K. For DML~\cite{kordopatis2017dml}, we obtain results for CC\_WEB\_VIDEO and FIVR200K from \cite{video-verification-fake-news}, and use the released source code\footnote{https://github.com/MKLab-ITI/ndvr-dml} for EVVE and FIVR5K, i.e. DML*. 
For TMK~\cite{baseline:tmk} and LAMV~\cite{Baraldi2018LAMVLT} on EVVE, we found discrepancies between their evaluation script and the original script from \cite{dataset:evve}. We provide corrected results, i.e. TMK*, LAMV*, using the original EVVE evaluation script~\cite{dataset:evve}. LAMV was not available in the released source code\footnote{\label{footnote:lamv}https://github.com/facebookresearch/videoalignment}, and we use TMK with frequency normalization (0.534 mAP on EVVE) as a close proxy for LAMV (0.536 mAP on EVVE~\cite{Baraldi2018LAMVLT}). We also evaluate TMK* and LAMV* on FIVR5K and FIVR200K. For other video-level methods, we report results from \cite{kordopatiszilos2019visil}.

\vspace{-3mm}
\paragraph{Frame-level Methods.} On EVVE, the authors of ViSiL~\cite{kordopatiszilos2019visil} were only able to download, process and evaluate on ~80\% of the dataset. We provide complete results for ViSiL using the released source code\footnote{https://github.com/MKLab-ITI/visil.git}, i.e. ViSiL$_\text{sym}$* and ViSiL$_{v}$*. For other frame-level methods, we report the updated results using deep network features in \cite{kordopatiszilos2019visil}.

\textbf{Near-duplicate Video Retrieval.}

In Table~\ref{tab:results:ccweb}, we compare our approach with other video-level and frame-level approaches on CC\_WEB\_VIDEO~\cite{dataset:cc-web-video}, and demonstrate that our approach achieves state-of-the-art performance for video-level NDVR. Similar to FIVR5K, we observe that our shot-level approach has intermediate performance to frame-level and video-level approaches. 
We also compare the run times for video-level VRAG and shot-level VRAG with ViSiL$_v$~\cite{kordopatiszilos2019visil} after extracting R-MAC features from each video frame. Video-level and shot-level VRAG takes 33mins and 52mins, respectively, while ViSiL uses 109mins to infer video similarities. 

\begin{table}[!htbp]
\small
\setlength\tabcolsep{3pt}
    \begin{tabular}{|l|l|c|c|c|c|}
        \hline \hline
        Type & Method & cc\_web & cc\_web* & cc\_web$_c$ & cc\_web$_c$* \\
        \hline \hline
        \multirow{3}{*}{Video} & \text{LBoW} & 0.957 & 0.906 & - & - \\
        & \text{DML} & \textbf{0.971} & 0.941 & 0.979 & 0.959 \\
        & \textbf{Ours} & \textbf{0.971} & \textbf{0.952} & \textbf{0.980} & \textbf{0.967} \\
        \hline \hline
        \multirow{2}{*}{Shot} & \textbf{Ours (CS)} & 0.975 & 0.955 & \textbf{0.987} & \textbf{0.977} \\
        & \textbf{Ours (SCS)} & \textbf{0.976} & \textbf{0.959} & 0.986 & \textbf{0.977} \\
        \hline\hline
        \multirow{6}{*}{Frame} & \text{CTE} & \textbf{0.996} & - & - & - \\
        & \text{DP} & 0.975 & 0.958 & 0.990 & 0.982 \\
        & \text{TN} & 0.978 & 0.965 & 0.991 & 0.987 \\
        & \text{ViSiL}$_f$ & 0.984 & 0.969 & 0.993 & 0.987 \\
        & \text{ViSiL}$_\text{sym}$ & 0.982 & 0.969 & 0.991 & 0.988 \\
        & \text{ViSiL}$_v$ & 0.985 & \textbf{0.971} & \textbf{0.996} & \textbf{0.993} \\
        \hline
    \end{tabular}
    \caption{Results on four different versions of CC\_WEB\_VIDEO. (*) denotes evaluation on the entire data set and the subscript $c$ uses cleaned annotations.} \vspace{-3mm}
    \label{tab:results:ccweb}
\end{table}

\begin{table*}[tb]
\setlength\tabcolsep{5pt}
\footnotesize
\centering
\begin{tabular}{|l||l||c||ccccccccccccc|}
\hline
Approach & Method & mAP & \multicolumn{13}{c|}{per event class} \\
\hline \hline
\multirow{9}{*}{Video}
    & LBoW$^\dagger$ & 0.469  & 0.323 & 0.373 & 0.062 & 0.392 & 0.306 & 0.232 & 0.205 & 0.127 & 0.060 & 0.376 & 0.233 & 0.769 & 0.713 \\    
    & \text{DML*} & 0.472  & 0.437 & 0.368 & 0.052 & 0.385 & 0.242 & 0.275 & 0.205 & 0.105 & 0.085 & 0.414 & 0.245 & 0.783 & 0.656 \\
    &\text{TMK*} & 0.469  & 0.508 & 0.306 & 0.139 & 0.366 & 0.294 & 0.244 & 0.208 & 0.125 & 0.152 & 0.287 & 0.213 & 0.810 & 0.614 \\
    & \text{LAMV*} & 0.493 & 0.649 & 0.321 & 0.157 & 0.411 & 0.319 & 0.241 & 0.224 & 0.124 & 0.194 & 0.257 & 0.191 & 0.857 & 0.660 \\
    & \text{LAMV+QE*} & 0.541 & 0.795 & 0.413 & \textbf{\underline{0.160}} & \underline{0.546} & \underline{0.376} & 0.297 & 0.235 & 0.124 & 0.236 & 0.257 & 0.185 & 0.907 & 0.754 \\
    & \text{LAMV} & 0.536 & 0.715 & 0.383 & 0.158 & 0.461 & 0.387 & 0.227 & 0.247 & 0.138 & 0.222 & 0.273 & 0.273 & 0.908 & 0.691 \\
    & \text{LAMV+QE} & 0.587 & 0.837 & 0.500 & 0.126 & \textbf{0.588} & \textbf{0.455} & \textbf{0.343} & \textbf{0.267} & 0.142 & 0.230 & 0.293 & 0.216 & \textbf{0.950} & 0.770 \\
    & \textbf{Ours} & 0.623 & 0.792  & 0.675 & 0.072 & 0.496 & 0.329 & 0.292 & \underline{0.256} & 0.241 & \underline{\textbf{0.497}} & 0.692 & 0.378 & 0.928 & 0.770 \\
    & \textbf{Ours+QE} & \textbf{\underline{0.653}}  & \textbf{\underline{0.888}} & \textbf{\underline{0.743}} & 0.042 & 0.505 & 0.342 & \underline{0.304} & 0.247 & \underline{\textbf{0.280}} & 0.489 & \underline{\textbf{0.782}} & \underline{\textbf{0.410}} & \underline{0.943} & \underline{\textbf{0.835}} \\
    \hline\hline
    \multirow{2}{*}{Shot} & \textbf{Ours (CS)} & 0.539 & 0.796 & 0.599 & 0.077 & \textbf{0.515} & 0.203 & \textbf{0.266} & 0.190 & 0.098 & 0.222 & 0.589 & 0.299 & 0.836 & \textbf{0.775} \\
    & \textbf{Ours (SCS)}  & \textbf{0.606} & \textbf{0.832} &\textbf{ 0.722} & \textbf{0.155} & 0.494 & \textbf{0.336} & 0.265 & \textbf{0.236} & \textbf{0.177} & \textbf{0.366} & \textbf{0.620} & \textbf{0.304} & \textbf{0.925} & 0.670 \\
    \hline\hline
    \multirow{3}{*}{Frame} & \text{ViSiL}$_{f}$ & 0.589 & 0.889 & 0.570 & 0.169 & 0.432 & 0.345 & \textbf{0.393} & \textbf{0.297} & 0.181 & 0.479 & 0.564 & 0.369 & 0.885 & 0.799 \\
    & \text{ViSiL}$_\text{sym}$* & 0.612 & \textbf{0.923} & \textbf{0.724} & \textbf{0.301} & 0.573 & \textbf{0.418} & 0.276 & 0.291 & \textbf{0.200} & \textbf{0.544} & 0.396 & 0.339 & \textbf{0.938} & 0.753 \\
    & \text{ViSiL}$_\text{v}$* & \textbf{0.618} & 0.920 & 0.713 & 0.222 & \textbf{0.589} & 0.350 & 0.345 & 0.276 & 0.169 & 0.444 & \textbf{0.567} & \textbf{0.375} & 0.909 &\textbf{0.842} \\
    \hline \hline
  \end{tabular}
  \caption{Comparison with state-of-the-art EVR approaches on EVVE. We use the same event class ordering as \cite{dataset:evve}. For video-level approaches, we also underline the result with the highest mAP, excluding the results with discrepancies in the evaluation script, i.e. LAMV and LAMV+QE. We report results obtained from the original EVVE evaluation script.} \vspace{-2mm}
  \label{tab:results:evve}
\end{table*}

\textbf{Fine-grained Incident Video Retrieval.} We evaluate on FIVR5K and FIVR200K in Table \ref{tab:results:fivr5k} and Table \ref{tab:results:fivr200k} respectively. Our results include the run times for available frame-level and video-level methods after extracting frame-level features. VRAG achieves state-of-the-art performance over video-level methods on FIVR datasets.
Generally, we observe a huge difference in retrieval performance between video-level and frame-level approaches on FIVR. Although our shot-level approach bridges the gap between video-level and frame-level while taking  $1.4\times$ longer than video-level VRAG, the dramatic disparity between the frame-level and video-level approaches warrants a qualitative inspection of FIVR200K.

\begin{table}[ht]
    \footnotesize
    \begin{tabular}{|l|l|c|c|c|c|}
        \hline \hline
        Type & Method & DSVR & CSVR & ISVR & Time\\
        \hline \hline
        \multirow{6}{*}{Video} & \text{LBoW$^\dagger$} & 0.351 & 0.320 & 0.298 & 4m 9s \\
        & \text{DML*} & 0.354 & 0.351 & 0.331 & 3m 29s \\
        & \text{TMK*} & 0.411 & 0.416 & 0.388  & 19m 9s \\ 
        & \text{LAMV*} & 0.498 & 0.488 & 0.426 & 20m 24s \\
        & \textbf{Ours} & \textbf{0.532} & \textbf{0.548} & \textbf{0.518} & 8m 27s \\
        \hline \hline
       \multirow{2}{*}{Shot} & \textbf{Ours (CS)} & \textbf{0.709} & \textbf{0.704} & \textbf{0.636}  &  \multirow{2}{*}{10m 35s}\\
        & \textbf{Ours (SCS)} & 0.596 & 0.606 & 0.575 &  \\
        \hline\hline
        \multirow{3}{*}{Frame} 
        & \text{ViSiL}$_{f}$ & 0.838 & 0.832 & 0.739 & - \\
        & \text{ViSiL}$_\text{sym}$ & 0.830 & 0.823 & 0.731 & 45m 24s \\
        & \text{ViSiL}$_v$ & \textbf{0.880} & \textbf{0.869} & \textbf{0.777} & 46m 39s  \\
        \hline
    \end{tabular}
    \caption{Results on FIVR5K} \vspace{-3mm}
    \label{tab:results:fivr5k}
\end{table}
\vspace{-3mm}
\begin{table}[ht]
    \footnotesize
    \begin{tabular}{|l|l|c|c|c|c|}
        \hline \hline
        Type & Method & DSVR & CSVR & ISVR & Time\\
        \hline \hline
        \multirow{6}{*}{Video} & \text{LBoW} & 0.378 & 0.361 & 0.297 & 3h 50m \\
        & \text{DML} & 0.398 & 0.351 & 0.331 & 3h 31m \\
        & \text{TMK*} & 0.417 & 0.394 & 0.319  & 18h 51m \\ 
        & \text{LAMV*} & \textbf{0.489} & 0.459 & 0.364 & 20h 30m \\
        & \textbf{Ours} & 0.484 & \textbf{0.470} & \textbf{0.399} & 6h 50m \\
        \hline \hline
       \multirow{2}{*}{Shot} & \textbf{Ours (CS)} & \textbf{0.723} & \textbf{0.678} & \textbf{0.554} & \multirow{2}{*}{9h 34m} \\
        & \textbf{Ours (SCS)} & 0.536 & 0.504 & 0.422 &  \\
        \hline\hline
        \multirow{5}{*}{Frame}
        & \text{DP} & 0.775 & 0.740 & 0.632 & - \\
        & \text{TN} & 0.724 & 0.699 & 0.589 & - \\
        & \text{ViSiL}$_{f}$ & 0.843 & 0.797 & 0.660 & - \\
        & \text{ViSiL}$_\text{sym}$ & 0.830 & 0.823 & 0.731 & 63h 31m \\
        & \text{ViSiL}$_{v}$ & \textbf{0.880} & \textbf{0.869} & \textbf{0.777} & 66h 18m\\
        \hline \hline
    \end{tabular}
    \caption{Results on FIVR200K} \vspace{-2mm}
    \label{tab:results:fivr200k}
\end{table}

We found that most FIVR200K queries comprise of single shots while their Duplicate Scene (DS) videos are sequences of visually diverse shots. Additionally, most shots in the DS videos have low visual and semantic correspondence with the query segment. Consequently, video-level methods are likely to yield noisy representations that are distinct to the query. On the other hand, higher fidelity approaches, i.e. frame-level and shot-level retrieval, that segment the video into multiple representations can preserve the query video segment and obtain more accurate video similarities. We show examples in the supplementary material.

\subsubsection{Event Video Retrieval}
In Table~\ref{tab:results:evve}, we evaluate our performance on EVVE~\cite{dataset:evve}. Without test augmentations, i.e. Query Expansion, VRAG outperforms state-of-the-art video-level and frame-level methods. We apply Average Query Expansion~\cite{query-expansion} on VRAG, i.e. VRAG+QE and it demonstrates further performance gains. Our shot-level approach is competitive with video-level and frame-level approaches and use 75mins of evaluation time. In contrast, video-level VRAG takes approximately 12mins for inference on EVVE, while ViSiL~\cite{kordopatiszilos2019visil} uses 18hrs. 
We attribute the difference in efficiency to paradigm differences between video-level and frame-level approaches: Our video-level method requires $620 + 2375 = 2995$ forward passes through VRAG to encode $n$ video embeddings. In contrast, ViSiL directly outputs the similarity between every query-candidate video pair. This requires $620 \times 2375 \approx 1.5\times10^6$ forward passes through ViSiL, which is orders of magnitude larger than the number of videos $n=2995$.


\section{Conclusion}

In this work, we introduce Video Region Attention Graph Network (VRAG) that utilizes self-attention during region aggregation and region pooling to generate video-level embeddings for efficient video retrieval. 
Specifically, we represent videos at a finer granularity through region-level features and encode video spatio-temporal dynamics through region-level relations. 
Our VRAG improves the state-of-the-art performance of video-level methods across multiple video retrieval datasets. We also introduce an intermediate approach to video-level and frame-level video retrieval that utilizes shots for video retrieval. We demonstrate that our shot-level approach bridges the gap in performance between video-level and frame-level approaches, where it obtains higher video retrieval performance at marginal increase in computational costs over video-level approaches.

\bibliography{aaai22}

\begin{thebibliography}{30}
\providecommand{\natexlab}[1]{#1}

\bibitem[{Baraldi et~al.(2018)Baraldi, Douze, Cucchiara, and
  J{\'e}gou}]{Baraldi2018LAMVLT}
Baraldi, L.; Douze, M.; Cucchiara, R.; and J{\'e}gou, H. 2018.
\newblock LAMV: Learning to Align and Match Videos with Kernelized Temporal
  Layers.
\newblock \emph{2018 IEEE/CVF Conference on Computer Vision and Pattern
  Recognition}, 7804--7813.

\bibitem[{Cai and Wang(2020)}]{gcn:over-smoothing1}
Cai, C.; and Wang, Y. 2020.
\newblock A Note on Over-Smoothing for Graph Neural Networks.
\newblock arXiv:2006.13318.

\bibitem[{Chen et~al.(2019)Chen, Lin, Li, Li, Zhou, and
  Sun}]{chen2019measuring}
Chen, D.; Lin, Y.; Li, W.; Li, P.; Zhou, J.; and Sun, X. 2019.
\newblock Measuring and Relieving the Over-smoothing Problem for Graph Neural
  Networks from the Topological View.
\newblock arXiv:1909.03211.

\bibitem[{{Chou}, {Chen}, and {Lee}(2015)}]{baseline:dp}
{Chou}, C.; {Chen}, H.; and {Lee}, S. 2015.
\newblock Pattern-Based Near-Duplicate Video Retrieval and Localization on
  Web-Scale Videos.
\newblock \emph{IEEE Transactions on Multimedia}, 17(3): 382--395.

\bibitem[{Clevert, Unterthiner, and Hochreiter(2016)}]{nonlinearity:elu}
Clevert, D.-A.; Unterthiner, T.; and Hochreiter, S. 2016.
\newblock Fast and Accurate Deep Network Learning by Exponential Linear Units
  (ELUs).
\newblock \emph{CoRR}, abs/1511.07289.

\bibitem[{{Douze} et~al.(2013){Douze}, {Revaud}, {Schmid}, and
  {Jégou}}]{query-expansion}
{Douze}, M.; {Revaud}, J.; {Schmid}, C.; and {Jégou}, H. 2013.
\newblock Stable Hyper-pooling and Query Expansion for Event Detection.
\newblock In \emph{2013 IEEE International Conference on Computer Vision},
  1825--1832.

\bibitem[{Fajtl et~al.(2018)Fajtl, Sokeh, Argyriou, Monekosso, and
  Remagnino}]{vid-summary:1812-01969}
Fajtl, J.; Sokeh, H.~S.; Argyriou, V.; Monekosso, D.; and Remagnino, P. 2018.
\newblock Summarizing Videos with Attention.
\newblock \emph{CoRR}, abs/1812.01969.

\bibitem[{Gabeur et~al.(2020)Gabeur, Sun, Alahari, and Schmid}]{mm-transformer}
Gabeur, V.; Sun, C.; Alahari, K.; and Schmid, C. 2020.
\newblock {Multi-modal Transformer for Video Retrieval}.
\newblock In \emph{{European Conference on Computer Vision (ECCV)}}.

\bibitem[{He et~al.(2015)He, Zhang, Ren, and Sun}]{resnet}
He, K.; Zhang, X.; Ren, S.; and Sun, J. 2015.
\newblock Deep Residual Learning for Image Recognition.
\newblock \emph{CoRR}, abs/1512.03385.

\bibitem[{Ji et~al.(2017)Ji, Xiong, Pang, and Li}]{vid-summary:1708-09545}
Ji, Z.; Xiong, K.; Pang, Y.; and Li, X. 2017.
\newblock Video Summarization with Attention-Based Encoder-Decoder Networks.
\newblock \emph{CoRR}, abs/1708.09545.

\bibitem[{Jiang, Jiang, and Wang(2014)}]{dataset:vcdb}
Jiang, Y.-G.; Jiang, Y.; and Wang, J. 2014.
\newblock VCDB: A Large-Scale Database for Partial Copy Detection in Videos.
\newblock In Fleet, D.; Pajdla, T.; Schiele, B.; and Tuytelaars, T., eds.,
  \emph{Computer Vision -- ECCV 2014}, 357--371. Cham: Springer International
  Publishing.
\newblock ISBN 978-3-319-10593-2.

\bibitem[{Kingma and Ba(2014)}]{optimizer:adam}
Kingma, D.; and Ba, J. 2014.
\newblock Adam: A Method for Stochastic Optimization.
\newblock \emph{International Conference on Learning Representations}.

\bibitem[{Kipf and Welling(2017)}]{gcn}
Kipf, T.~N.; and Welling, M. 2017.
\newblock {Semi-Supervised Classification with Graph Convolutional Networks}.
\newblock In \emph{Proceedings of the 5th International Conference on Learning
  Representations}, ICLR '17.

\bibitem[{Kordopatis{-}Zilos et~al.(2019)Kordopatis{-}Zilos, Papadopoulos,
  Patras, and Kompatsiaris}]{video-verification-fake-news}
Kordopatis{-}Zilos, G.; Papadopoulos, S.; Patras, I.; and Kompatsiaris, I.
  2019.
\newblock Finding Near-Duplicate Videos in Large-Scale Collections.
\newblock In Mezaris, V.; Nixon, L. J.~B.; Papadopoulos, S.; and Teyssou, D.,
  eds., \emph{Video Verification in the Fake News Era}, 91--126. Springer.

\bibitem[{Kordopatis-Zilos et~al.(2019)Kordopatis-Zilos, Papadopoulos, Patras,
  and Kompatsiaris}]{dataset:fivr200k}
Kordopatis-Zilos, G.; Papadopoulos, S.; Patras, I.; and Kompatsiaris, I. 2019.
\newblock FIVR: Fine-grained Incident Video Retrieval.
\newblock \emph{IEEE Transactions on Multimedia}.

\bibitem[{Kordopatis-Zilos et~al.(2017{\natexlab{a}})Kordopatis-Zilos,
  Papadopoulos, Patras, and Kompatsiaris}]{lbow}
Kordopatis-Zilos, G.; Papadopoulos, S.; Patras, I.; and Kompatsiaris, Y.
  2017{\natexlab{a}}.
\newblock Near-Duplicate Video Retrieval by Aggregating Intermediate CNN
  Layers.
\newblock In \emph{MMM}.

\bibitem[{Kordopatis-Zilos et~al.(2017{\natexlab{b}})Kordopatis-Zilos,
  Papadopoulos, Patras, and Kompatsiaris}]{kordopatis2017dml}
Kordopatis-Zilos, G.; Papadopoulos, S.; Patras, I.; and Kompatsiaris, Y.
  2017{\natexlab{b}}.
\newblock Near-Duplicate Video Retrieval with Deep Metric Learning.
\newblock In \emph{2017 IEEE International Conference on Computer Vision
  Workshop (ICCVW)}.

\bibitem[{{Kordopatis-Zilos} et~al.(2019){Kordopatis-Zilos}, {Papadopoulos},
  {Patras}, and {Kompatsiaris}}]{kordopatiszilos2019visil}
{Kordopatis-Zilos}, G.; {Papadopoulos}, S.; {Patras}, I.; and {Kompatsiaris},
  Y. 2019.
\newblock ViSiL: Fine-Grained Spatio-Temporal Video Similarity Learning.
\newblock In \emph{2019 IEEE/CVF International Conference on Computer Vision
  (ICCV)}, 6350--6359.

\bibitem[{Liu et~al.(2019)Liu, Albanie, Nagrani, and
  Zisserman}]{collaborative-experts}
Liu, Y.; Albanie, S.; Nagrani, A.; and Zisserman, A. 2019.
\newblock Use What You Have: Video retrieval using representations from
  collaborative experts.
\newblock In \emph{BMVC}.

\bibitem[{Miech, Laptev, and Sivic(2018)}]{moEE}
Miech, A.; Laptev, I.; and Sivic, J. 2018.
\newblock Learning a Text-Video Embedding from Incomplete and Heterogeneous
  Data.
\newblock \emph{CoRR}, abs/1804.02516.

\bibitem[{{Miech} et~al.(2019){Miech}, {Zhukov}, {Alayrac}, {Tapaswi},
  {Laptev}, and {Sivic}}]{howto100m}
{Miech}, A.; {Zhukov}, D.; {Alayrac}, J.; {Tapaswi}, M.; {Laptev}, I.; and
  {Sivic}, J. 2019.
\newblock HowTo100M: Learning a Text-Video Embedding by Watching Hundred
  Million Narrated Video Clips.
\newblock In \emph{2019 IEEE/CVF International Conference on Computer Vision
  (ICCV)}, 2630--2640.

\bibitem[{Poullot et~al.(2015)Poullot, Tsukatani, Phuong~Nguyen, J\'{e}gou, and
  Satoh}]{baseline:tmk}
Poullot, S.; Tsukatani, S.; Phuong~Nguyen, A.; J\'{e}gou, H.; and Satoh, S.
  2015.
\newblock Temporal Matching Kernel with Explicit Feature Maps.
\newblock In \emph{Proceedings of the 23rd ACM International Conference on
  Multimedia}, MM '15, 381–390. New York, NY, USA: Association for Computing
  Machinery.
\newblock ISBN 9781450334594.

\bibitem[{{Revaud} et~al.(2013){Revaud}, {Douze}, {Schmid}, and
  {Jégou}}]{dataset:evve}
{Revaud}, J.; {Douze}, M.; {Schmid}, C.; and {Jégou}, H. 2013.
\newblock Event Retrieval in Large Video Collections with Circulant Temporal
  Encoding.
\newblock In \emph{2013 IEEE Conference on Computer Vision and Pattern
  Recognition}, 2459--2466.

\bibitem[{Tan et~al.(2009)Tan, Ngo, Hong, and Chua}]{temporal_network}
Tan, H.-K.; Ngo, C.-W.; Hong, R.; and Chua, T.-S. 2009.
\newblock Scalable Detection of Partial Near-Duplicate Videos by
  Visual-Temporal Consistency.
\newblock In \emph{Proceedings of the 17th ACM International Conference on
  Multimedia}, MM '09, 145–154. New York, NY, USA: Association for Computing
  Machinery.
\newblock ISBN 9781605586083.

\bibitem[{Tolias, Sicre, and J{\'e}gou(2016)}]{rmac}
Tolias, G.; Sicre, R.; and J{\'e}gou, H. 2016.
\newblock Particular object retrieval with integral max-pooling of CNN
  activations.
\newblock \emph{CoRR}, abs/1511.05879.

\bibitem[{Vassileios~Balntas and Mikolajczyk(2016)}]{triplet-margin-loss}
Vassileios~Balntas, D.~P., Edgar~Riba; and Mikolajczyk, K. 2016.
\newblock Learning local feature descriptors with triplets and shallow
  convolutional neural networks.
\newblock In Richard C.~Wilson, E. R.~H.; and Smith, W. A.~P., eds.,
  \emph{Proceedings of the British Machine Vision Conference (BMVC)},
  119.1--119.11. BMVA Press.
\newblock ISBN 1-901725-59-6.

\bibitem[{Vaswani et~al.(2017)Vaswani, Shazeer, Parmar, Uszkoreit, Jones,
  Gomez, Kaiser, and Polosukhin}]{attention-is-all-you-need}
Vaswani, A.; Shazeer, N.; Parmar, N.; Uszkoreit, J.; Jones, L.; Gomez, A.~N.;
  Kaiser, L.; and Polosukhin, I. 2017.
\newblock Attention Is All You Need.
\newblock \emph{CoRR}, abs/1706.03762.

\bibitem[{Veličković et~al.(2018)Veličković, Cucurull, Casanova, Romero,
  Liò, and Bengio}]{graph-attention}
Veličković, P.; Cucurull, G.; Casanova, A.; Romero, A.; Liò, P.; and Bengio,
  Y. 2018.
\newblock Graph Attention Networks.
\newblock arXiv:1710.10903.

\bibitem[{Wu, Hauptmann, and Ngo(2007)}]{dataset:cc-web-video}
Wu, X.; Hauptmann, A.~G.; and Ngo, C.-W. 2007.
\newblock Practical Elimination of Near-Duplicates from Web Video Search.
\newblock In \emph{Proceedings of the 15th ACM International Conference on
  Multimedia}, MM '07, 218–227. New York, NY, USA: Association for Computing
  Machinery.
\newblock ISBN 9781595937025.

\bibitem[{Zhou, Qiao, and Xiang(2018)}]{vid-summary:zhou2018deep}
Zhou, K.; Qiao, Y.; and Xiang, T. 2018.
\newblock Deep Reinforcement Learning for Unsupervised Video Summarization with
  Diversity-Representativeness Reward.
\newblock arXiv:1801.00054.

\end{thebibliography}


\begin{thebibliography}{4}
\providecommand{\natexlab}[1]{#1}

\bibitem[{Kordopatis-Zilos et~al.(2019)Kordopatis-Zilos, Papadopoulos, Patras,
  and Kompatsiaris}]{dataset:fivr200k}
Kordopatis-Zilos, G.; Papadopoulos, S.; Patras, I.; and Kompatsiaris, I. 2019.
\newblock FIVR: Fine-grained Incident Video Retrieval.
\newblock \emph{IEEE Transactions on Multimedia}.

\bibitem[{Kordopatis-Zilos et~al.(2017)Kordopatis-Zilos, Papadopoulos, Patras,
  and Kompatsiaris}]{kordopatis2017dml}
Kordopatis-Zilos, G.; Papadopoulos, S.; Patras, I.; and Kompatsiaris, Y. 2017.
\newblock Near-Duplicate Video Retrieval with Deep Metric Learning.
\newblock In \emph{2017 IEEE International Conference on Computer Vision
  Workshop (ICCVW)}.

\bibitem[{{Kordopatis-Zilos} et~al.(2019){Kordopatis-Zilos}, {Papadopoulos},
  {Patras}, and {Kompatsiaris}}]{kordopatiszilos2019visil}
{Kordopatis-Zilos}, G.; {Papadopoulos}, S.; {Patras}, I.; and {Kompatsiaris},
  Y. 2019.
\newblock ViSiL: Fine-Grained Spatio-Temporal Video Similarity Learning.
\newblock In \emph{2019 IEEE/CVF International Conference on Computer Vision
  (ICCV)}, 6350--6359.

\bibitem[{{Revaud} et~al.(2013){Revaud}, {Douze}, {Schmid}, and
  {Jégou}}]{dataset:evve}
{Revaud}, J.; {Douze}, M.; {Schmid}, C.; and {Jégou}, H. 2013.
\newblock Event Retrieval in Large Video Collections with Circulant Temporal
  Encoding.
\newblock In \emph{2013 IEEE Conference on Computer Vision and Pattern
  Recognition}, 2459--2466.

\end{thebibliography}

\clearpage

\end{document}


\title{Supplementary Material for \\ VRAG: Region Attention Graphs for Content-Based Video Retrieval}

\maketitle

\section{Qualitative Results}

\subsection{T-SNE Embeddings}

\begin{figure}[h]
    \centering
    \includegraphics[width=1\columnwidth]{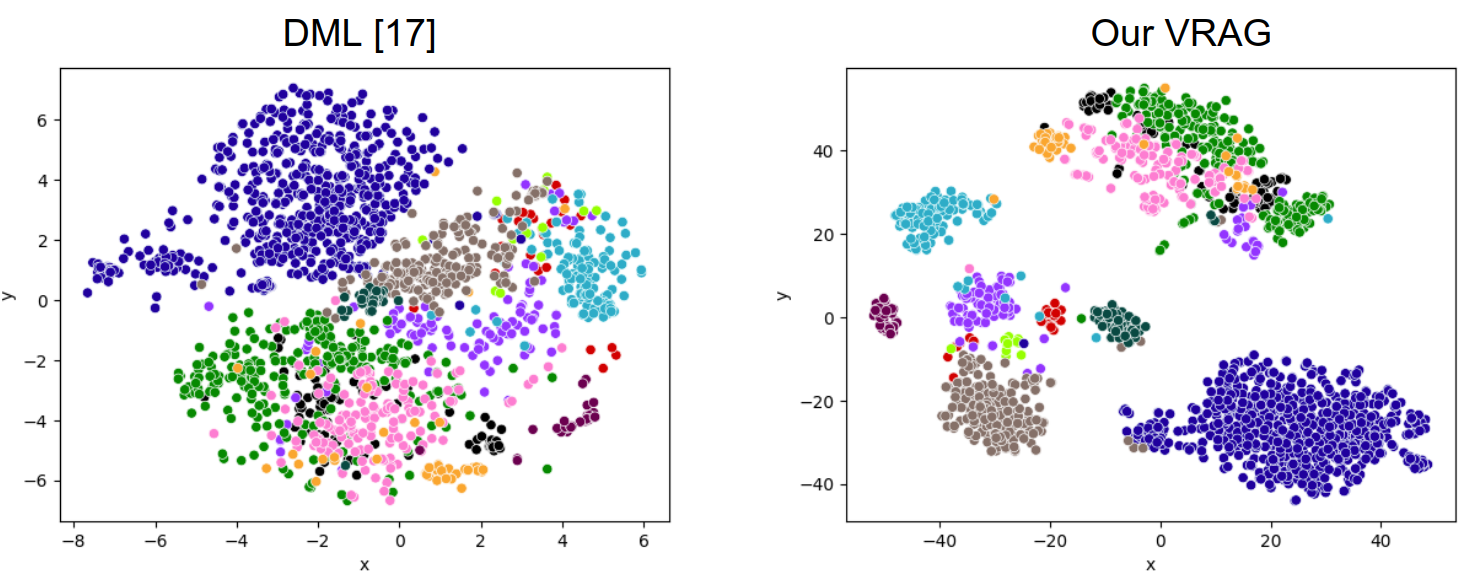}
    \caption{T-SNE embedding visualizations on EVVE. (Left) DML~\cite{kordopatis2017dml}. (Right) Our VRAG.}
    \label{fig:tsne-evve}
\end{figure}

\begin{figure}[h]
    \centering
    \includegraphics[width=1\columnwidth]{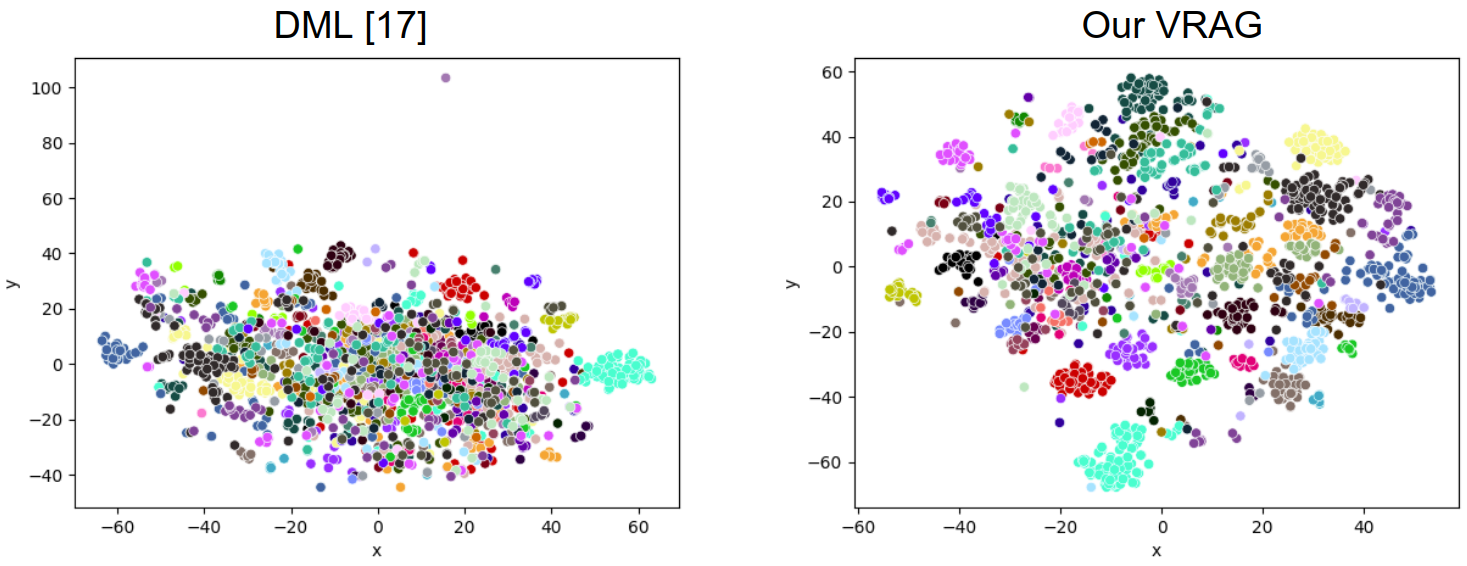}
    \caption{T-SNE embedding visualizations on FIVR-5K. (Left) DML~\cite{kordopatis2017dml}. (Right) Our VRAG.}
    \label{fig:tsne-fivr5k}
\end{figure}

In this section, we visualize our video-level representations into T-SNE embeddings for EVVE~\cite{dataset:evve} in Figure~\ref{fig:tsne-evve} and FIVR5K~\cite{dataset:fivr200k, kordopatiszilos2019visil} in Figure~\ref{fig:tsne-fivr5k}.

Across both datasets, we observe that our VRAG forms more visible and well-separated T-SNE embedding clusters compared to the existing video-level method DML~\cite{kordopatis2017dml}.

\subsection{Retrieval Results}

\begin{figure*}[h]
    \centering
    \includegraphics[width=1.5\columnwidth]{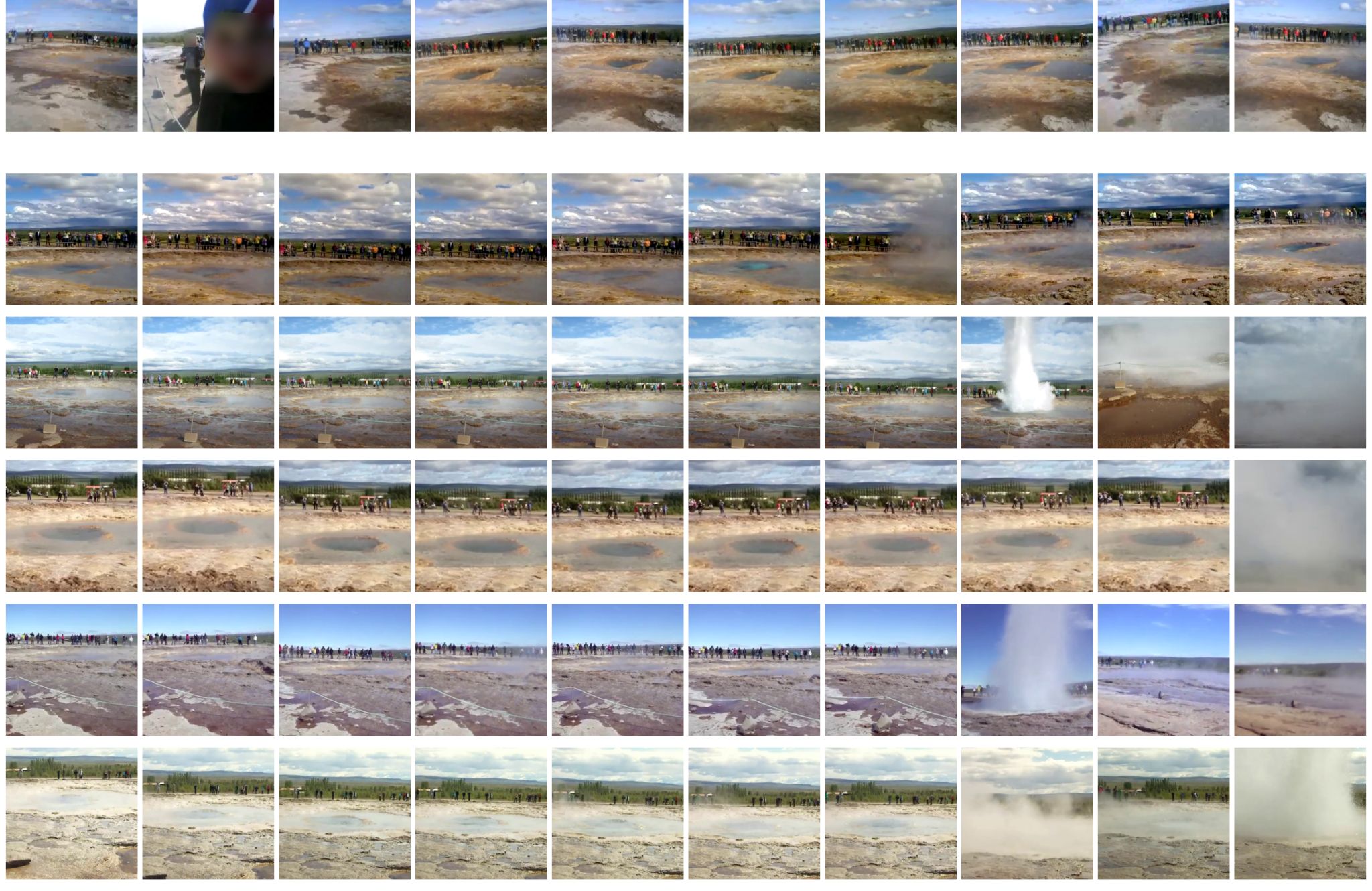}
    \caption{Positive geyser event retrieval example.}
    \label{fig:pos:geyser}
\end{figure*}

\begin{figure*}[h]
    \centering
    \includegraphics[width=1.5\columnwidth]{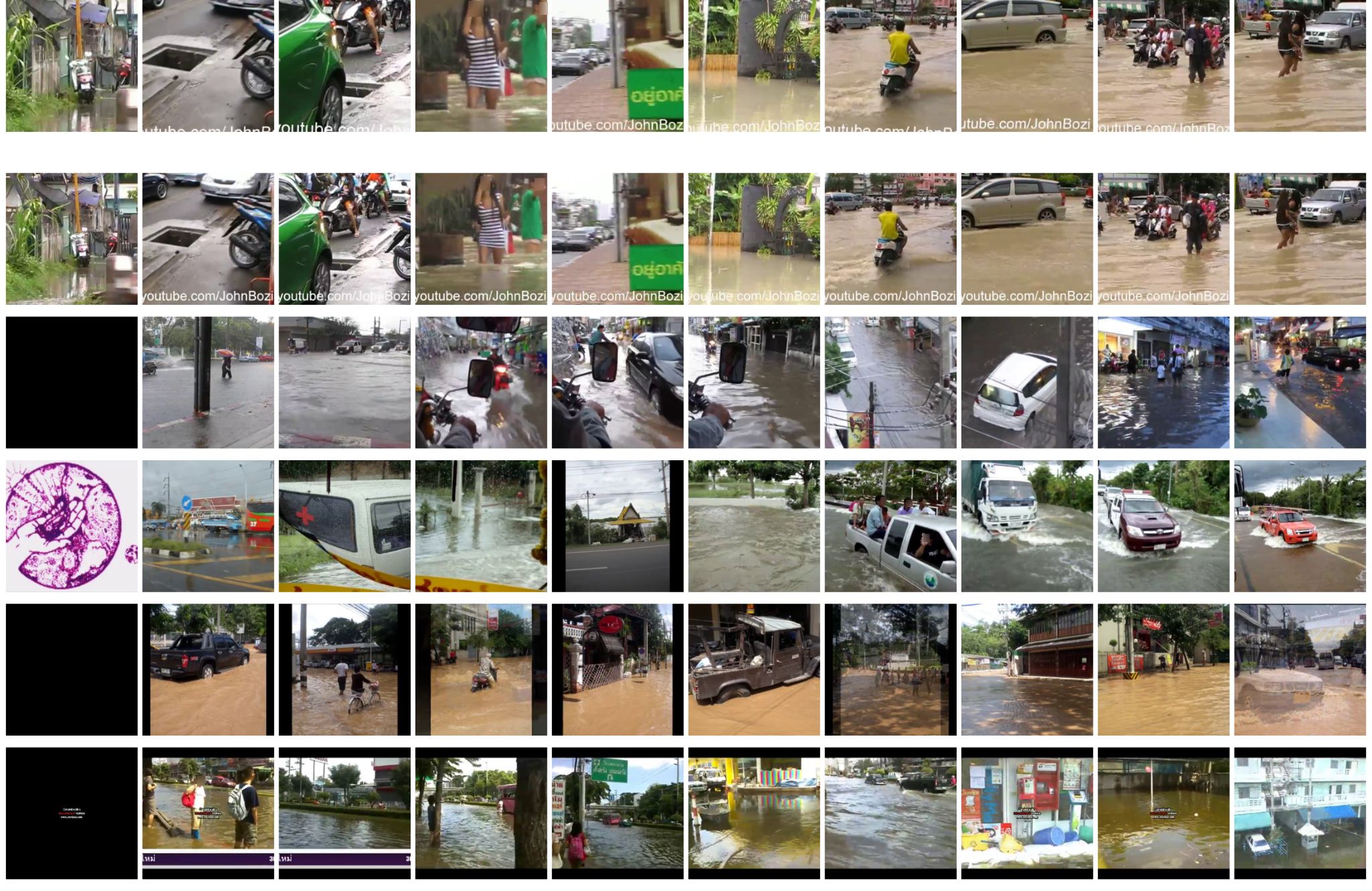}
    \caption{Positive flooding event retrieval example.}
    \label{fig:pos:flood}
\end{figure*}

\begin{figure*}[h]
    \centering
    \includegraphics[width=1.5\columnwidth]{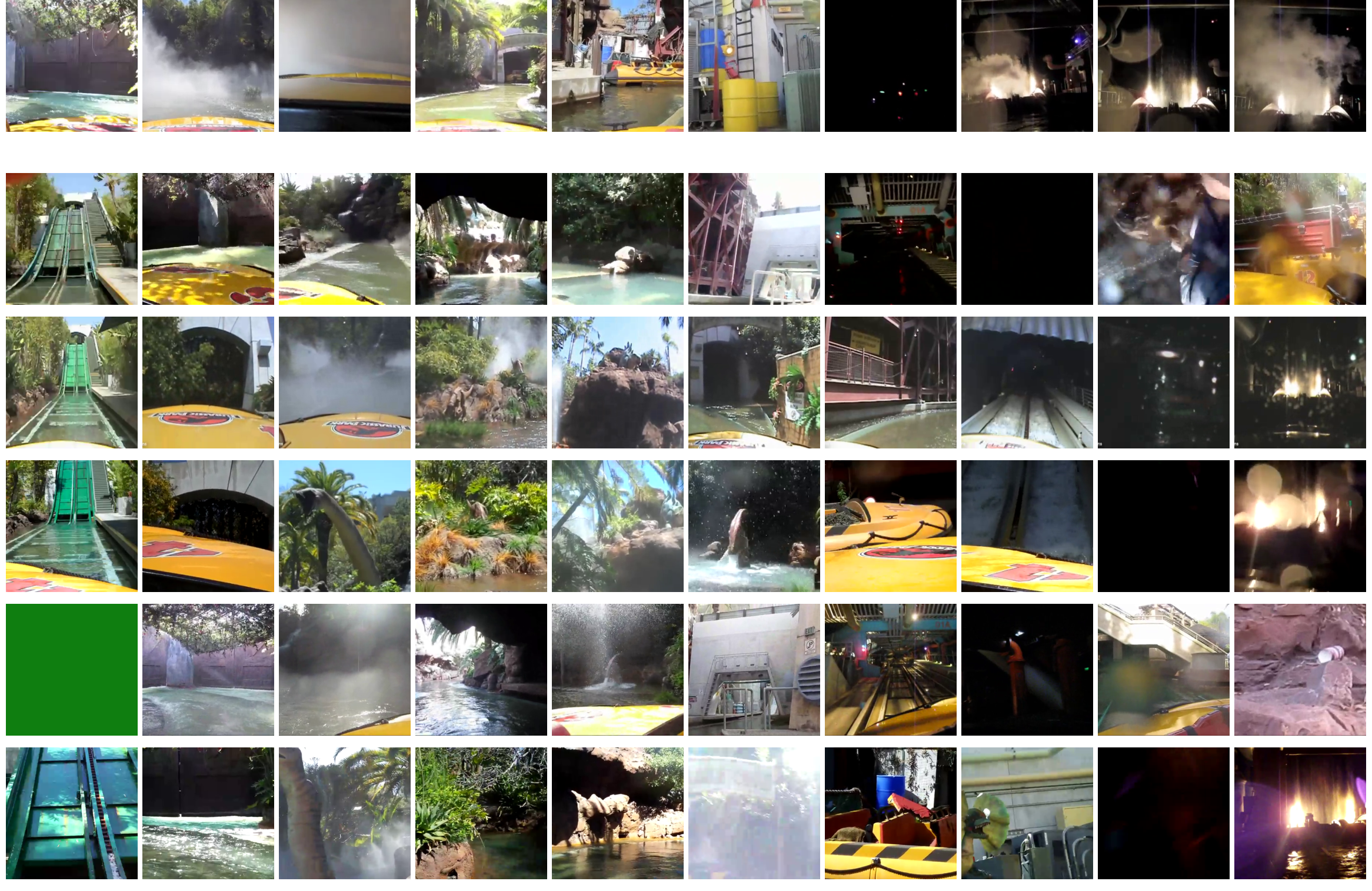}
    \caption{Positive water park event retrieval example.}
    \label{fig:pos:water-park}
\end{figure*}

\begin{figure*}[h]
    \centering
    \includegraphics[width=1.5\columnwidth]{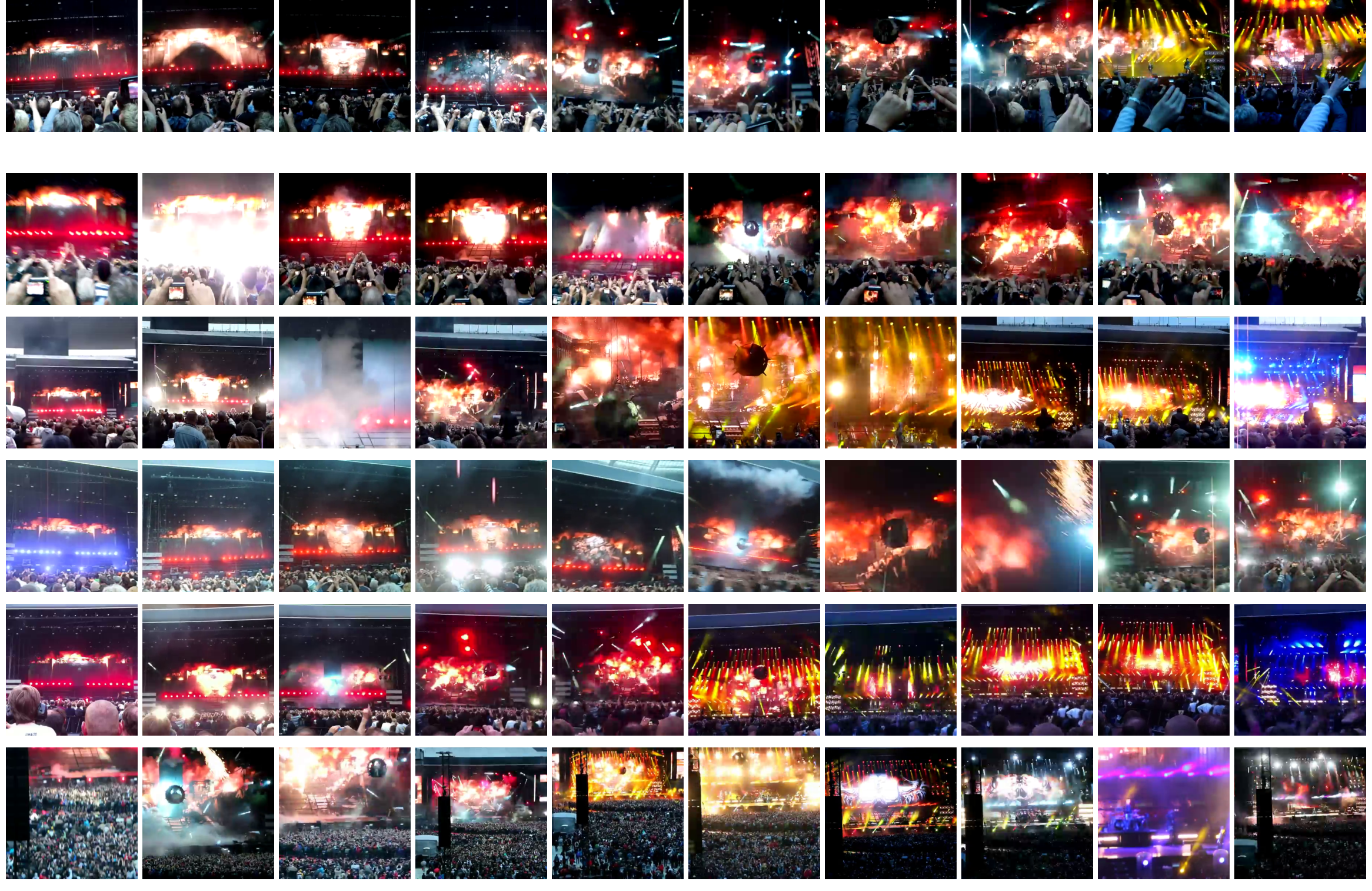}
    \caption{Positive live performance retrieval example.}
    \label{fig:pos:performance}
\end{figure*}

Additionally, we include examples of retrieval results on the EVVE~\cite{dataset:evve} dataset. In our retrieval examples, we visualize the five closest database videos to the query videos. We demonstrate positive results in Figures~{\ref{fig:pos:geyser} to \ref{fig:pos:performance}}. From Figures~\ref{fig:pos:flood} and \ref{fig:pos:performance}, we see that our VRAG can retrieve videos of similar events taken from different viewpoints. In Figures~\ref{fig:pos:geyser} and \ref{fig:pos:water-park}, VRAG retrieves videos of the same event taken from different points in time.

\begin{figure*}[h]
    \centering
    \includegraphics[width=1.5\columnwidth]{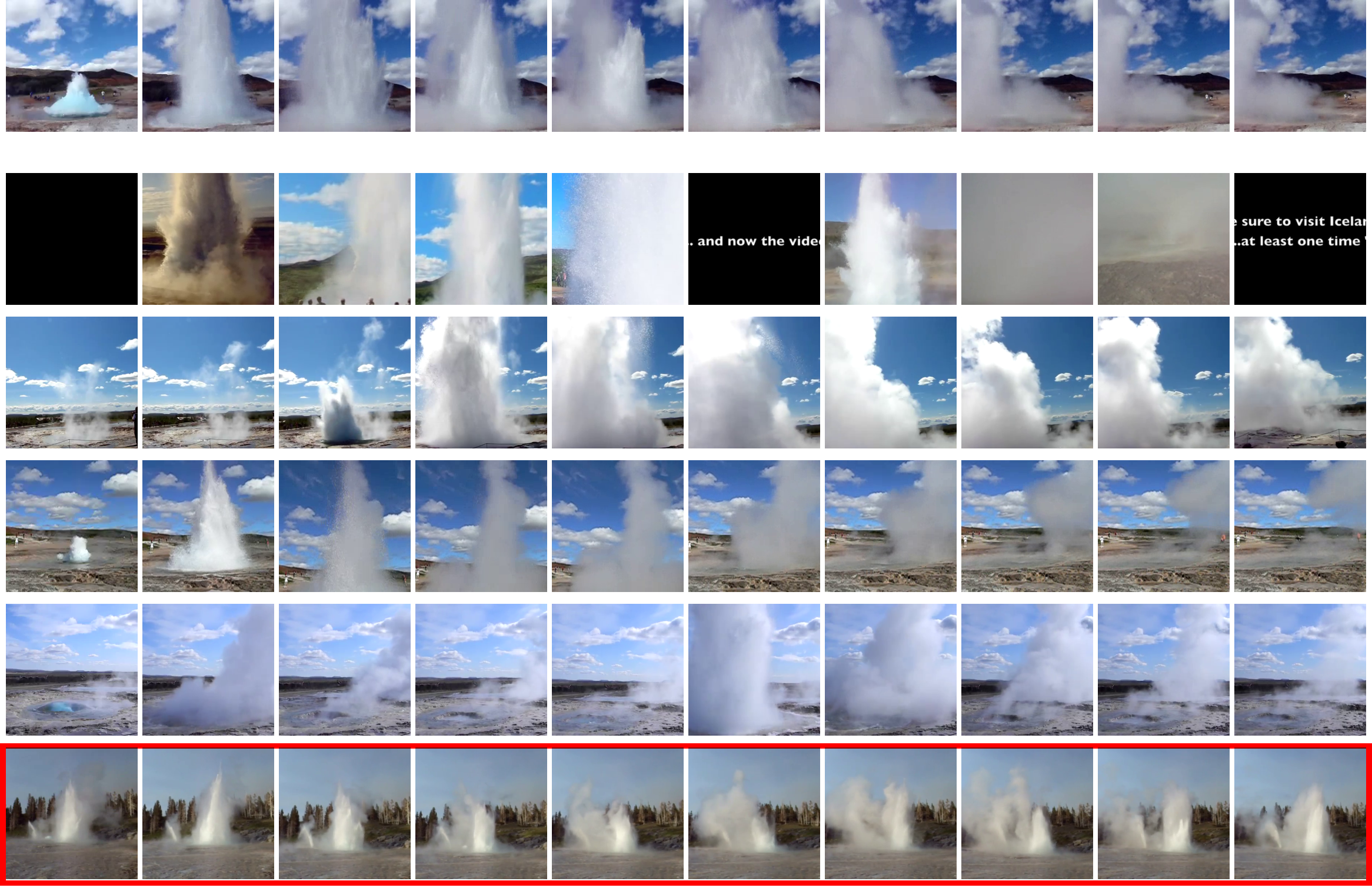}
    \caption{Negative geyser event retrieval example. Negative videos have red borders.}
    \label{fig:neg:geyser}
\end{figure*}

\begin{figure*}[h]
    \centering
    \includegraphics[width=1.5\columnwidth]{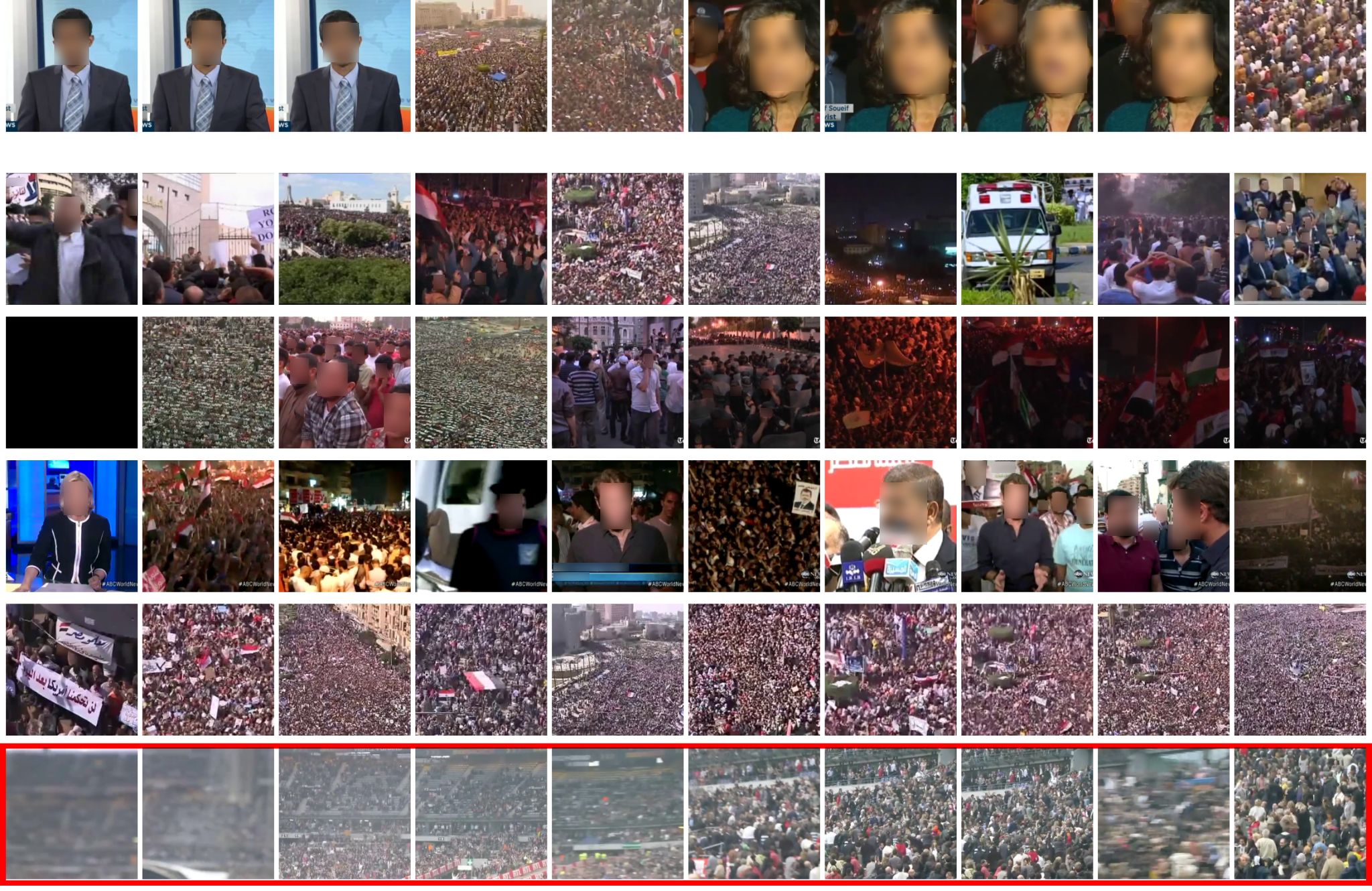}
    \caption{Negative protests event retrieval example.  Negative videos have red borders.}
    \label{fig:neg:protest}
\end{figure*}

\begin{figure*}[h]
    \centering
    \includegraphics[width=1.5\columnwidth]{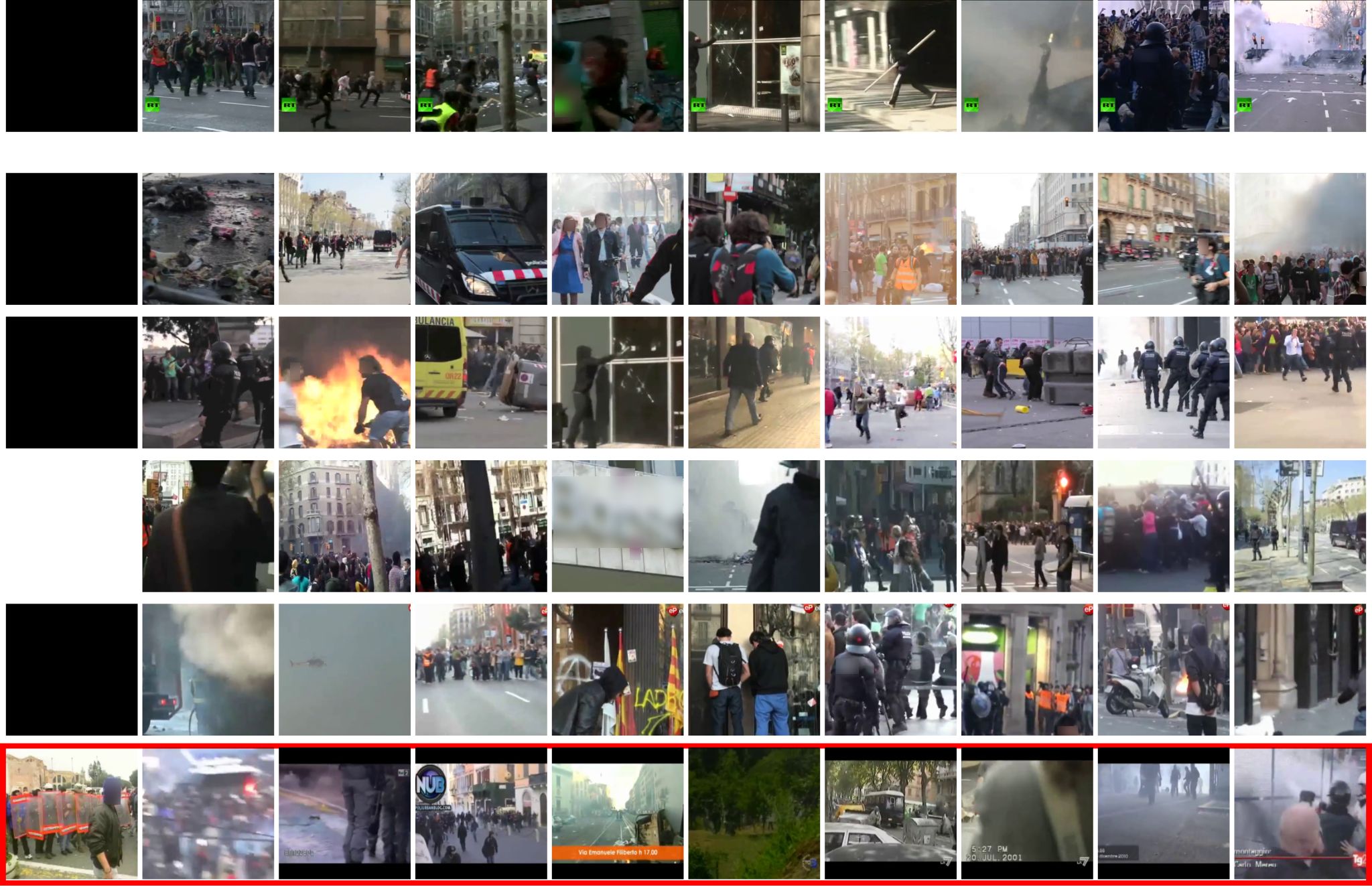}
    \caption{Negative riots event retrieval example. Negative videos have red borders.}
    \label{fig:neg:riot}
\end{figure*}

\begin{figure*}[h]
    \centering
    \includegraphics[width=1.5\columnwidth]{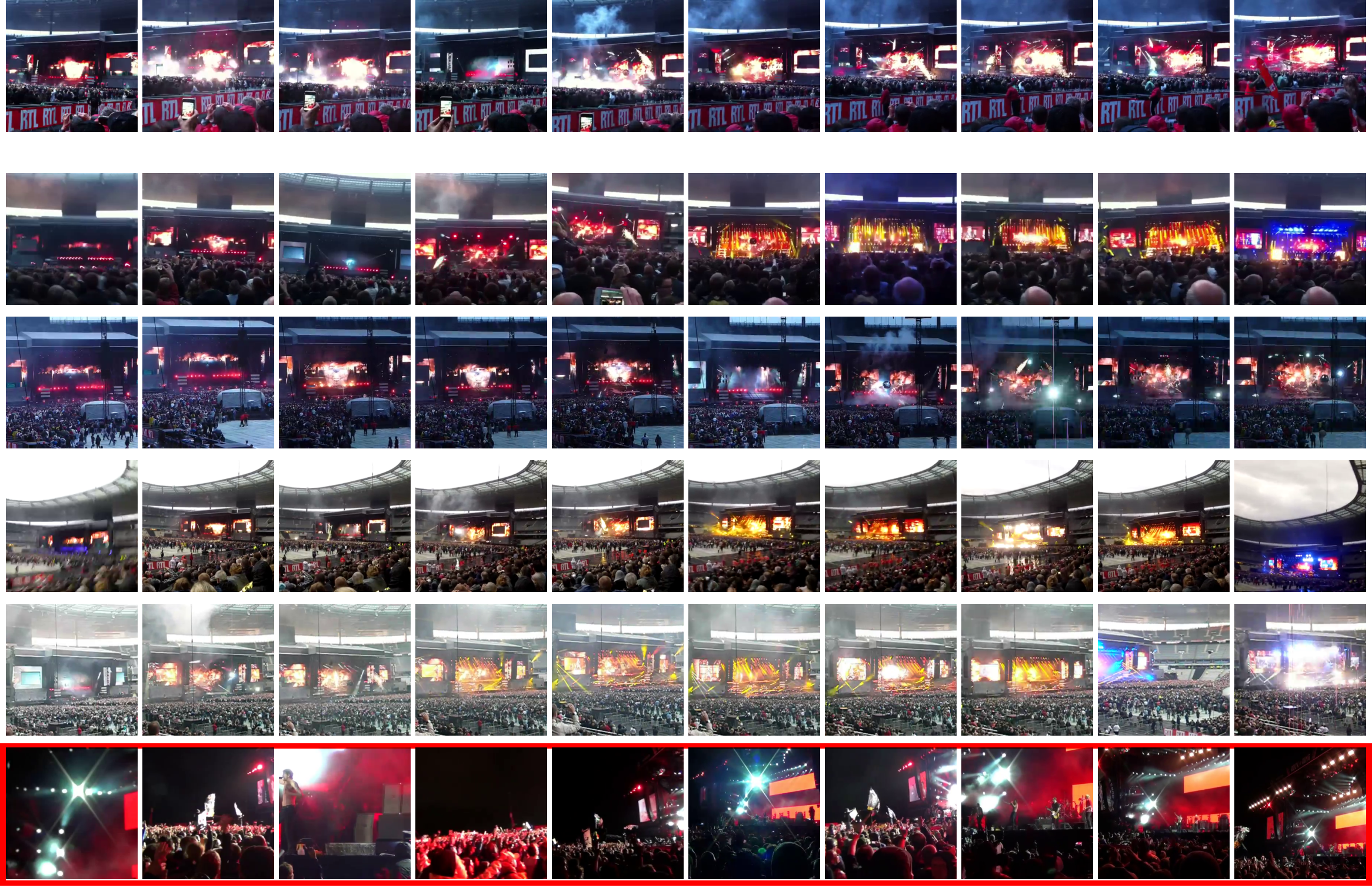}
    \caption{Negative live performance retrieval example. Negative videos have red borders.}
    \label{fig:neg:performance}
\end{figure*}

Figures~\ref{fig:neg:geyser} to \ref{fig:neg:performance} demonstrate some failure cases of our method with negatively-related videos marked in red borders. In Figures~{\ref{fig:neg:geyser}, \ref{fig:neg:riot} and \ref{fig:neg:performance}}, we see that our VRAG pulls negative videos that are semantically related to the query video. Specifically, Figure~\ref{fig:neg:geyser} includes a negative video from a different geyser, Figure~\ref{fig:neg:riot} has a negative video from another riot, and Figure~\ref{fig:pos:performance} shows a negative video from a separate live performance. However, our VRAG may confuse videos 
in instances where distractor videos contain similar elements to query videos as positives. For example in Figure~\ref{fig:neg:protest}, VRAG retrieves a negative video depicting a crowded sports stadium as a close positive to a query video of a protest with multiple scenes of crowds.

\bibliography{aaai22}